\documentclass{article}
\usepackage{ijcai26}

\usepackage{times}
\usepackage{soul}
\usepackage{url}
\usepackage[hidelinks]{hyperref}
\usepackage[utf8]{inputenc}
\usepackage[small]{caption}
\usepackage{graphicx}
\usepackage{amsmath}
\usepackage{amsthm}
\usepackage{booktabs}
\usepackage{algorithm}
\usepackage{algorithmic}
\usepackage[switch]{lineno}

\usepackage{multirow, multicol}
\usepackage{makecell}
\usepackage{ragged2e}
\usepackage{amsfonts}
\usepackage{xspace}
\usepackage{wrapfig}
\usepackage{subcaption} 
\usepackage{amssymb}

\makeatletter
\DeclareRobustCommand\onedot{\futurelet\@let@token\@onedot}
\def\@onedot{\ifx\@let@token.\else.\null\fi\xspace}

\def\eg{\emph{e.g}\onedot} 
\def\ie{\emph{i.e}\onedot}

\makeatother

\title{Mitigating Backdoors via Decoy Shortcuts and Knowledge Decoupling}

\author{
Zixuan Zhu$^{1,2}$
\and
Rui Wang$^{1,2}$\thanks{Corresponding author} \and
Lihua Jing$^{1,2}$\And
Jinwen Zhong$^{1,2}$\\
\affiliations
$^1$Institute of Information Engineering, Chinese Academy of Sciences\\
$^2$School of Cyber Security, University of Chinese Academy of Sciences\\
\emails
\{zhuzixuan, wangrui, jinglihua, zhongjinwen\}@iie.ac.cn
}

\begin{document}

\maketitle

\begin{abstract}
    Backdoor attacks pose a serious threat to deep neural networks, especially when training relies on third-party data, allowing adversaries to inject malicious behaviors through data poisoning. In this work, we reveal that backdoor behaviors tend to be absorbed by a simpler parallel branch when jointly trained with the main network. Motivated by this insight, we propose \emph{Trapping and Removing (TR)}, a simple yet effective training-time defense that introduces a lightweight shortcut branch as a “honeypot” to trap backdoor knowledge. After training, backdoors can be removed by discarding the shortcut, without requiring any additional data. To further enhance backdoor isolation while maintaining benign performance, we design a knowledge decoupling strategy with entropy-based weight assignment, encouraging poisoned samples to flow through the honeypot while guiding the main network to focus on benign learning. In addition, we introduce an automatic shortcut generation strategy to improve generalization across model architectures. Extensive experiments on four benchmark datasets and five model architectures demonstrate that our approach effectively mitigates a wide range of backdoor attacks while preserving performance on benign data. Code: \href{https://github.com/Zixuan-Zhu/TR}{github.com/Zixuan-Zhu/TR}.

\end{abstract}
\section{Introduction}
\label{sec:intro}


With large-scale training data and powerful computational resources, deep neural networks (DNNs) have achieved remarkable performance and are widely deployed in real-world applications~\cite{ResNet,wang2020pillar,zhang2020unsupervised}. However, the high cost of collecting training data often drives developers to rely on third-party datasets, inadvertently exposing models to backdoor threats.



Backdoor attackers can manipulate models by poisoning a small subset of training data, embedding a specific \emph{trigger} and assigning a corresponding \emph{target label}. During training, the model implicitly learns this malicious association, causing any input containing the trigger to be misclassified as the target label at inference time. Deploying backdoored models can have severe consequences, especially in safety-critical applications such as autonomous driving. Prior work indicates that backdoors, once implanted, are difficult to erase or unlearn~\cite{ESTI}, making training-time countermeasures particularly important.

\begin{figure}[t]
  \centering 
  \includegraphics[width=\linewidth]{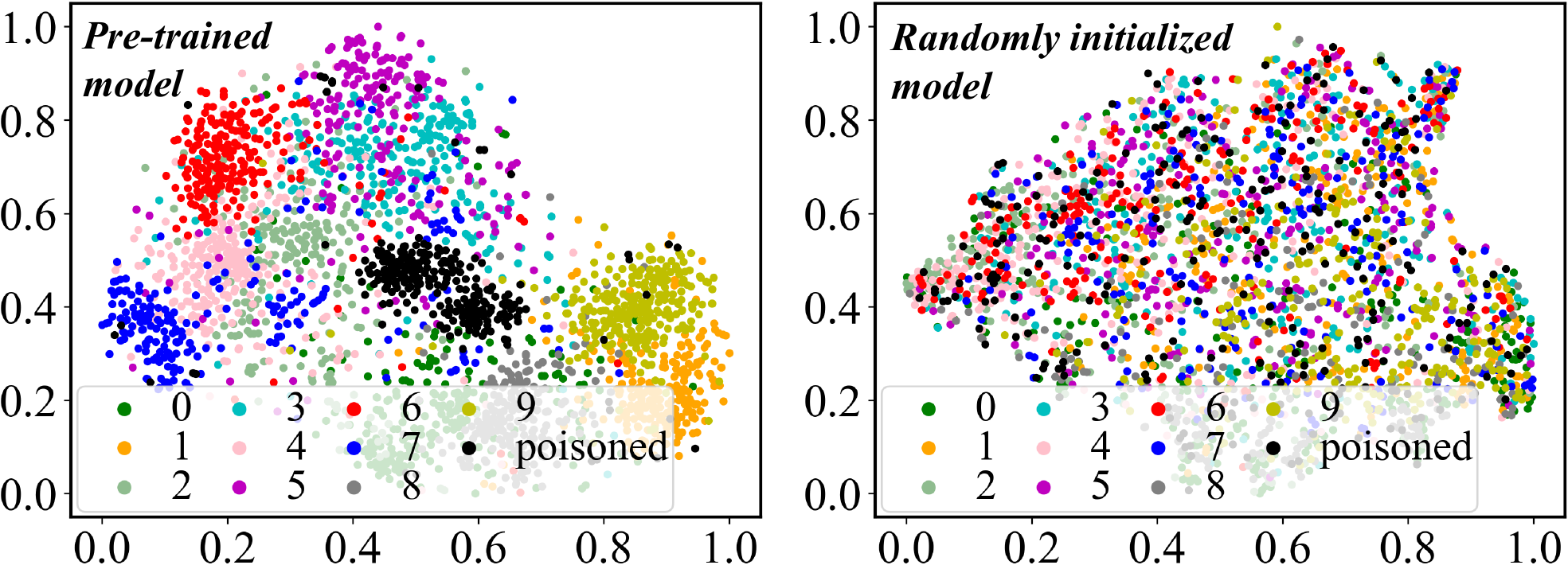}
  \caption{t-SNE visualizations of poisoned samples in the feature space produced by ResNet-18 models: pre-trained on ImageNet (left) vs. randomly initialized (right).}
  \label{fig: feature_vis_pretrain_or_scratch}
\end{figure}

PBE~\cite{PBE} and Neural Cleanse~\cite{neuralclense} show that backdoors tend to form shortcut connections in the model, directly associating trigger patterns with target labels. These methods exploit this property to detect and erase backdoors after training. Inspired by this, we are motivated to explore \emph{whether a decoy shortcut can be explicitly introduced during training to trap backdoor knowledge as it forms}.  Several studies~\cite{setting,liu2023shortcuts} explore backdoor defense in NLP by confining backdoor knowledge within shallow layers or lightweight models, thereby preventing the main network from internalizing it. These approaches are primarily designed for scenarios where attackers fine-tune pre-trained clean models with poisoned data. In contrast, backdoor injection in image classification is typically performed during training from scratch. To illustrate this difference, Figure~\ref{fig: feature_vis_pretrain_or_scratch} compares feature distributions of poisoned samples crafted by BadNets~\cite{badnets} on CIFAR-10~\cite{CIFAR}, under both pre-trained and randomly initialized models.  In the left subfigure (pre-trained), poisoned samples cluster in a sparse and relatively isolated region, suggesting a degree of separation from benign ones. In contrast, in the right subfigure (from-scratch), poisoned and benign samples appear highly entangled, making separation more challenging and limiting the effectiveness of NLP-oriented defenses in this setting.

In our experiments, we find that backdoor knowledge can be readily captured by a simple parallel branch when training from scratch, as demonstrated in Section~\ref{sec: Distinctive Learning Behaviors of Two Branches for Backdoor Attacks}.
Building on this observation, we propose a novel defense mechanism, \emph{Trapping and Removing (TR)}, that protects networks from backdoor attacks by trapping all backdoor knowledge within a dedicated ``honeypot" branch. This honeypot is introduced prior to training and implemented as a lightweight shortcut running parallel to the original backbone. In early epochs, the shortcut can effectively capture backdoor knowledge, albeit along with a few benign knowledge. To better focus its learning on backdoor knowledge and direct the original network to solely learn benign knowledge in subsequent training, we design a knowledge decoupling strategy equipped with an entropy-based weight assignment module, which dynamically guides poisoned samples toward learning through the shortcut branch while benign samples are processed by the original network. In this manner, the backdoor can be effectively removed by discarding the shortcut after training without resorting to any additional data, while maintaining high performance on benign data. To facilitate practical deployment, we further introduce an adaptive shortcut generation strategy that automatically tailors the shortcut to the backbone network.


The main contributions of this paper are as follows: 
\begin{itemize}
	\item  We reveal an inherent characteristic of backdoor attacks: backdoor knowledge tends to be learned through a simpler path during the early stages of training.

	\item Building on this insight, we propose \emph{TR}, a training-time defense based on decoy shortcuts and knowledge decoupling, which enables clean model training under backdoor attacks without any additional data and generalizes to both training-from-scratch and fine-tuning attacks.

    \item We design an adaptive strategy to automatically generate shortcuts for backbone networks, enhancing generalization and facilitating practical deployment.
    
    \item We evaluate our method on multiple datasets and models against ten backdoor attacks, showing robust performance and outperforming ten state-of-the-art defenses.

\end{itemize}



\section{Related Work}
\label{sec:related_work}

\subsection{Backdoor Attack and Defense}
\paragraph{Backdoor Attack.}
Based on the attack strategy, poisoning-based backdoor attacks can be broadly classified into non-optimized and optimized attacks. \emph{Non-optimized attacks}~\cite{datafree,DUBA} craft poisoned samples before training and distribute them to victims, often via third-party datasets. Early methods~\cite{badnets,blend} used simple, visible triggers, while later works such as WaNet~\cite{wanet} and Dynamic~\cite{Dynamic} introduced imperceptible or sample-specific triggers. To evade manual inspection, clean-label variants~\cite{SIG,PCBA} embed triggers into benign samples under the target class. \emph{Optimized attacks}~\cite{IBA,Lotus} instead generate and refine poisoned samples during the victim model's training process. These methods often rely on additional losses or co-trained trigger generators to enhance stealth. IBA~\cite{IBA} explicitly encourages backdoor invisibility in feature space, while LOTUS~\cite{Lotus} leverages partition-specific training to control the backdoor behavior more precisely. Optimized attacks require control over the model training process, which is beyond the scope of this paper.



\paragraph{Backdoor Defense.}
Backdoor inhibiting~\cite{zhu2023victim,ESTI} and erasing~\cite{NAD,FP} are two widely used defense strategies. \emph{Inhibiting-based defenses} aim to suppress backdoor formation during training by modifying the training process. For example, CBD~\cite{CBD} applies causal theory to prevent learning the backdoor path, DBD~\cite{DBD} decouples the training process, and ASD~\cite{ASD} continuously divides the training data. \emph{Erasing-based defenses} focus on purifying poisoned models after backdoor training and typically require benign samples. FP~\cite{FP} prunes low-activation neurons and fine-tunes the model, while NAD~\cite{NAD} performs fine-tuning followed by knowledge distillation. The defense proposed in this paper falls into the inhibiting-based category.

\subsection{Honeypot and Shortcut in Backdoor Defense} 
Several works adopt similar concepts of honeypots or shortcuts for backdoor defense, which differ from our approach. SSFT~\cite{yang2023backdoor} erases backdoors by removing skip connections in ResNets and finetuning with 10\% benign samples. T\&R~\cite{wang2022trap} baits the backdoor into a classification head and replaces it with another head finetuned on benign samples. DPoE~\cite{liu2023shortcuts} and \cite{setting} target NLP tasks, using shallow layers to absorb backdoor knowledge while preventing deeper layers from learning it. DPoE further combines this with a Product-of-Experts framework to isolate clean behavior. However, in NLP, backdoor injection is typically implemented by fine-tuning pre-trained clean models with poisoned data, where benign and backdoor knowledge are naturally decoupled, making defense more manageable. In contrast, we address the more challenging training-from-scratch injection, where benign and backdoor knowledge are entangled. Besides, we introduce a parallel shortcut branch as the honeypot, which is effective not only against training-from-scratch attacks but also against fine-tuning attacks. 

\subsection{Knowledge Decoupling}
Feature decoupling \cite{wang2021decoupling,yang2022decoupling} is a representative of knowledge decoupling, aiming to separate feature representations into independent components, minimize redundancy, and improve the model’s ability to learn more discriminative information. For example, DeAOT \cite{yang2022decoupling} decouples object-agnostic and object-specific features for better propagating annotations in semi-supervised video object segmentation. FDTrack \cite{jin2023multi} applies feature decoupling to balance the conflicting feature requirements for detection and re-identification in multi-object tracking. The specific implementation and objectives of knowledge decoupling can vary depending on the task. In this paper, we propose to decouple the benign and poisoned knowledge during backdoor training for defending.

\begin{figure*}[t]
  \centering
  \begin{subfigure}[c]{0.49\textwidth}
    \centering
    \includegraphics[width=\textwidth]{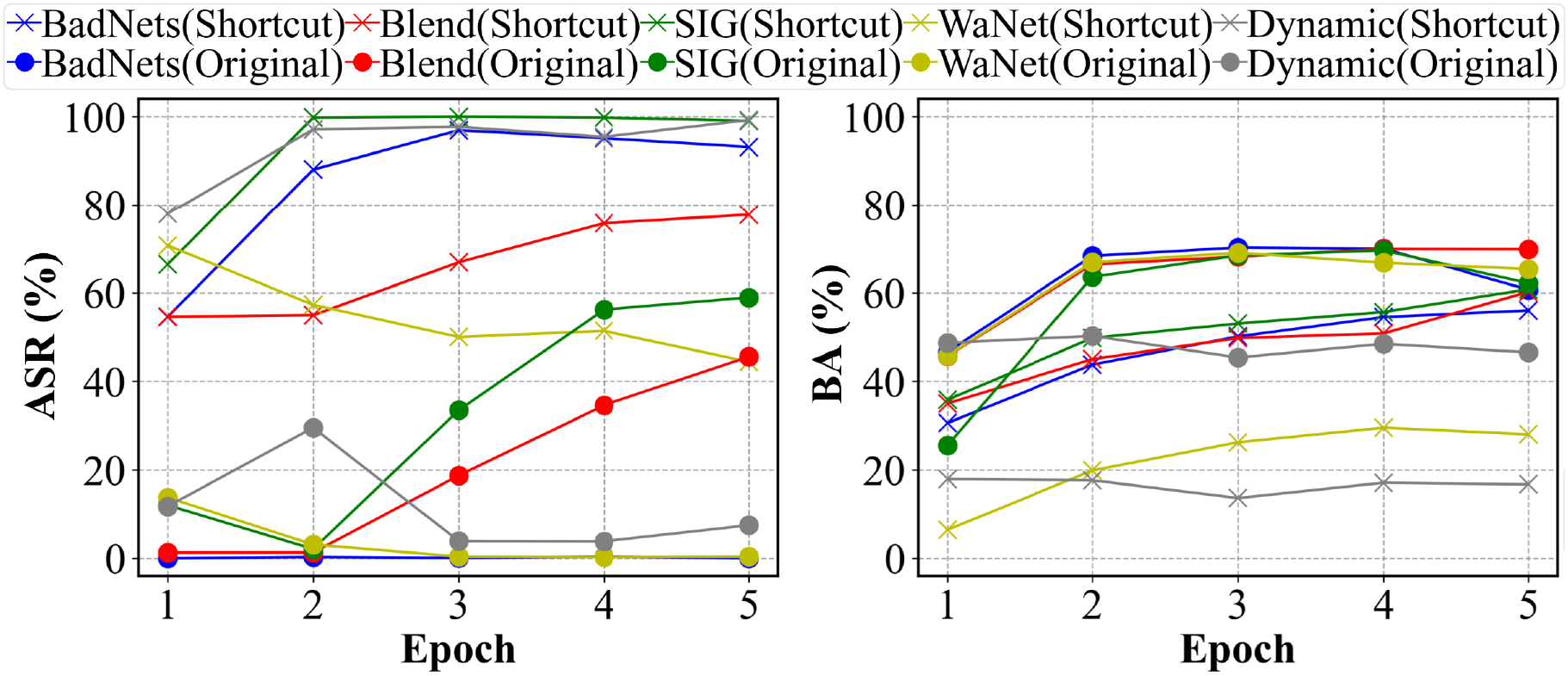}
    \caption{BA and ASR of two branches under five attacks}
    \label{fig: motivation}
  \end{subfigure}
  \hfill
  \begin{subfigure}[c]{0.49\textwidth}
    \centering
    \includegraphics[width=\textwidth]{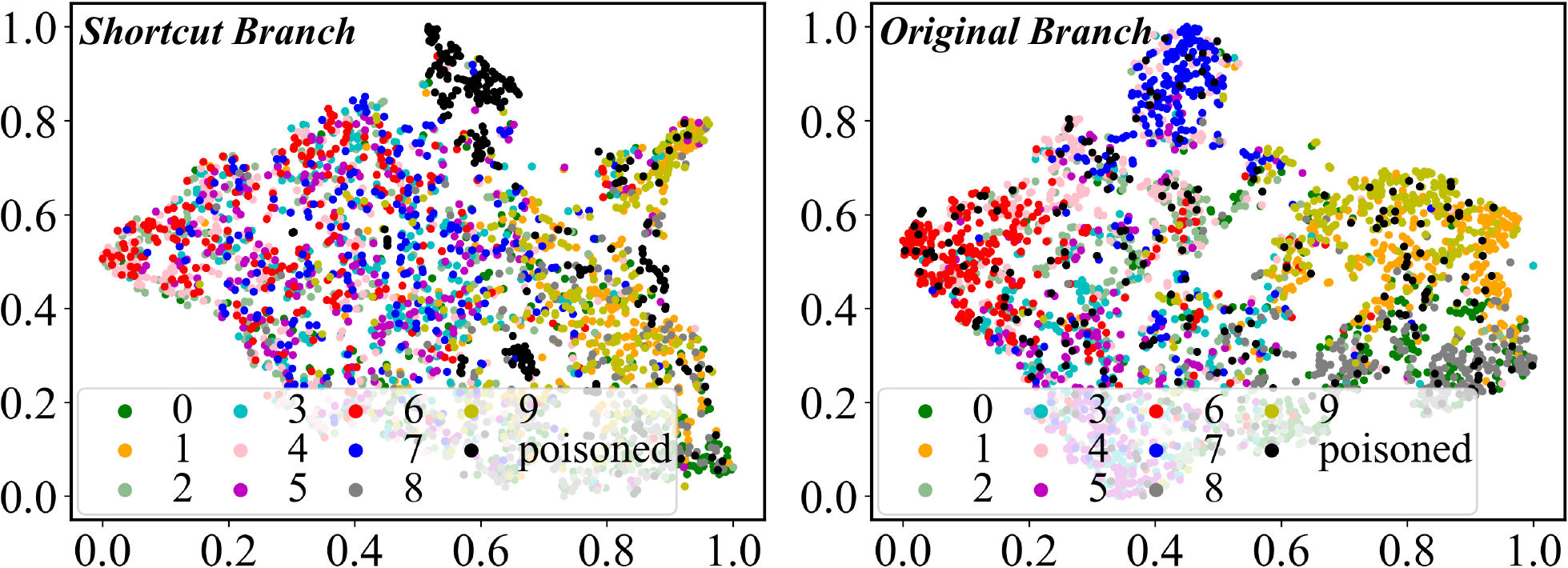}
    \caption{t-SNE visualization of poisoned samples in the feature spaces of two branches (epoch 1)}
    \label{fig: feature_vis_two_branches}
  \end{subfigure}

  \caption{Visualization of learning behaviors of original and shortcut branches under attacks.}
  \label{fig: two-subfigures}
\end{figure*}

\section{Preliminaries}
\subsection{Poisoning-based Backdoor Attacks}
We consider poisoning-based backdoor attacks in image classification. Let $\mathcal{D} = \{(\boldsymbol{x}_i, y_i)\}_{i=1}^N$ be the training set with $N$ samples, where each image $\boldsymbol{x}_i \in \{0, 1, \dots, 255\}^{C \times H \times W}$ and label $y_i \in \{0, \dots, K \mkern-5mu - \mkern-5mu 1\}$. Here, $K$ is the number of classes, and $H$, $W$, $C$ denote image height, width, and channels. In a backdoor attack, the attacker selects a subset of $\mathcal{D}$ and applies a modification function $G(\cdot)$ to generate poisoned samples: $\mathcal{D}_m = \{(\boldsymbol{x}^{\prime}, y_t) | \boldsymbol{x}^{\prime} = G(\boldsymbol{x}), (\boldsymbol{x}, y) \in \mathcal{D}\setminus \mathcal{D}_b \}$, where $\mathcal{D}_b$ contains the remaining benign samples and $y_t$ is the target label. The resulting poisoned dataset is $\mathcal{D}_p = \mathcal{D}_m \cup \mathcal{D}_b$, which victim users use unknowingly to train their models. At test time, the attacker triggers misclassification by applying $G(\cdot)$. The poisoning rate is defined as $r = \frac{|\mathcal{D}_m|}{|\mathcal{D}|}$.

\subsection{Threat Model}
\paragraph{Attacker’s Capability.} We consider a common threat scenario involving third-party datasets, where an attacker can arbitrarily modify the training data before releasing it to victim users, but has no access to information beyond the dataset itself, such as the model architecture or loss functions. The attacker’s objective is to cause the trained model to predict a predefined target label for triggered inputs while maintaining correct predictions on benign inputs. \emph{We further consider adaptive attack scenarios in Appendix~\ref{Appendix：defense against adaptive attacks}, where the attacker is aware of the existence of our defense}.

\paragraph{Defender's Capability.} The defender is assumed to have full control over the training process, but no prior knowledge of the backdoor and no access to any additional samples. Whether the training dataset is poisoned is also unknown to the defender. The defense objective is to prevent the trained model from predicting triggered samples as the target label while preserving accuracy on benign data.

\section{Method}
\label{sec:method}

\subsection{Distinct Learning Behaviors of Dual Branches under Backdoor Attacks}
\label{sec: Distinctive Learning Behaviors of Two Branches for Backdoor Attacks}
In this subsection, we analyze the distinct learning behaviors of the original and shortcut branches under backdoor attacks and discuss their inherent mechanisms.

\paragraph{Settings.}
We conduct experiments on CIFAR-10 using a modified WRN-16-1~\cite{WRN} architecture, augmented with a two-layer shortcut that directly connects the input to the classifier. During training, the final prediction is computed as the average of the outputs from both the shortcut and original branches (i.e., equal weights of 0.5). Figure~\ref{fig: motivation} reports the benign accuracy (BA) and attack success rate (ASR) of both branches under five different attacks during the early training epochs. In addition, Figure~\ref{fig: feature_vis_two_branches} visualizes the feature distributions of BadNets-poisoned samples produced by the two branches after a single epoch. More detailed experimental settings can be found in \emph{Appendix}~\ref{Appendix: Detailed Settings for Learning Behaviors Analysis}.

\paragraph{Results.}

As shown in Figure~\ref{fig: motivation}, the shortcut branch exhibits higher ASR but lower BA than the original branch during early training. This disparity arises from intrinsic differences between benign and poisoned samples: benign samples contain complex, diverse feature patterns better captured by the deeper original branch, whereas poisoned samples share similar trigger patterns and require only a few neurons to associate with the target label~\cite{FP}. Our theoretical analysis (\emph{Appendix}~\ref{Appendix: perspective of gradient}) further confirms that this structural asymmetry induces a gradient focusing effect ($||\nabla_{\theta_h}|| \gg ||\nabla_{\theta_o}||$) for poisoned samples. This mechanism results in a pronounced simplicity bias, where the shortcut establishes a dominant gradient flow to preferentially capture backdoor knowledge. Figure~\ref{fig: feature_vis_two_branches} further corroborates this: in the shortcut branch's feature space, most poisoned samples form a tight cluster while clean samples are relatively scattered; conversely, in the original branch's space, clean samples of the same class form coherent clusters, and poisoned samples are more dispersed—often near their ground-truth classes. These results suggest that backdoor knowledge tends to be learned by the simpler shortcut, while benign knowledge is primarily captured by the more complex original branch.

\begin{figure*}[t]
\begin{center}
\includegraphics[width=0.95\linewidth]{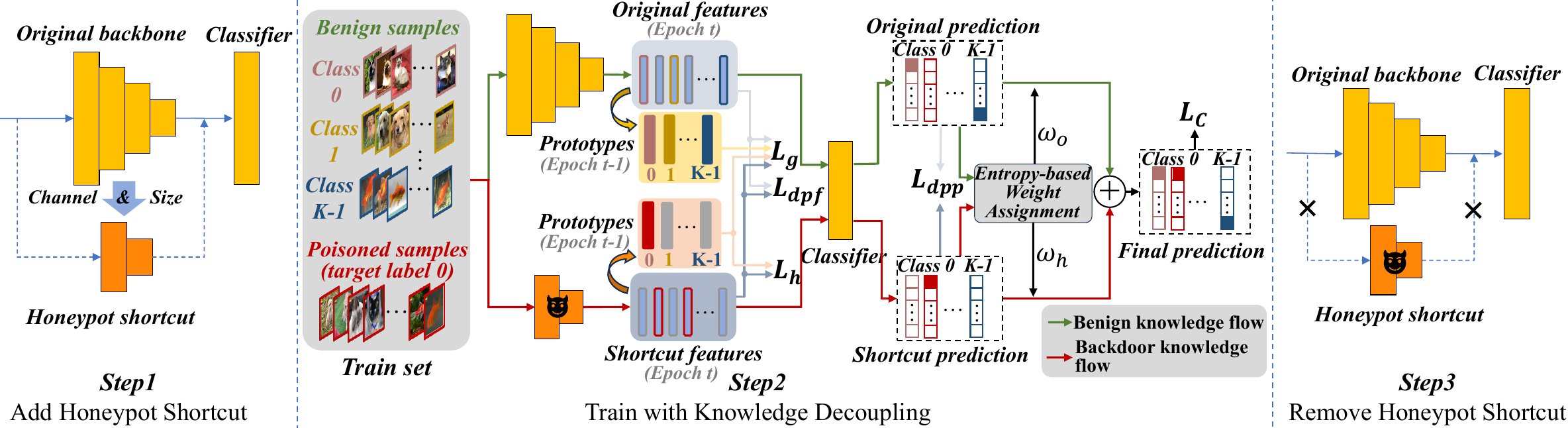}
\end{center}
   \caption{Our defense pipeline comprises three steps. First, we add a simple honeypot shortcut that runs parallel to the original backbone. Next, we train the model on the poisoned dataset while decoupling the knowledge between the original backbone and the honeypot shortcut. This enables the honeypot to capture backdoor knowledge while the original backbone retains only benign knowledge. Finally, after training, we erase the backdoor by removing the honeypot shortcut from the model.}
\label{fig: overview}
\end{figure*}


\subsection{Adaptive Shortcut Generation}
\label{sec: Honeypot Shortcut Designing}
As Figure~\ref{fig: overview} illustrates, our defense method consists of three steps. First, we add a simple honeypot shortcut that runs parallel to the original backbone before training. The modified network is denoted as $\{S_o, S_h, S_c\}$, where $S_o$ is the original backbone, $S_h$ is the honeypot shortcut, and $S_c$ is the classifier. Next, we train the modified network on the poisoned dataset, carefully decoupling the knowledge between the original branch $\{S_o, S_c\}$ and the shortcut branch $\{S_h, S_c\}$, enabling the shortcut to stabilize with backdoor-related information while allowing the original network to retain only benign knowledge. Finally, the backdoor can be easily erased by simply removing the shortcut from the model, without affecting the original network's benign performance.

We construct the honeypot shortcut as a direct path from the input to the final classifier. Specifically, each $2\times$ downsampling operation in the backbone is replaced with a $3 \times 3$ convolution of stride 2, while matching the corresponding channel dimensions. For instance, two convolution blocks suffice for WRN-16-1, which has two downsampling operations, while ResNet-18 requires three. More details on shortcut designs for CNNs and ViTs are provided in \emph{Appendix}~\ref{Appendix: Adaptive Shortcut Designing}.

To help the shortcut capture backdoor knowledge more quickly and effectively, we incorporate a spatial attention mechanism to highlight key regions. Let $f \in \mathbb{R}^{h \times w \times c}$ be the feature map, where $h$, $w$, and $c$ denote height, width, and channels. Following~\cite{cbam}, the attention map $M \in \mathbb{R}^{h \times\ w}$ is computed by applying a convolution to $f$ with a single output channel.
We then employ the attention map to highlight informative regions of the feature maps:
\begin{equation}
     f^{\prime} = ReLU(M) \otimes f,
\label{attention}
\end{equation}
where $\otimes$ indicates element-wise multiplication and $f^{\prime}$ represents the output of the attention module. 

For poisoned samples, spatial attention can quickly focus on trigger patterns due to their similarity and strong correlation with the target label. Consequently, each convolution block in the shortcut is structured as \emph{Conv-Attention-BN-ReLU} to enhance its ability to capture backdoor knowledge.


\subsection{Entropy-based Weight Assignment}
In the modified network, we define the final prediction $P(\boldsymbol{x})$ as a weighted combination of the outputs from both two branches: 
\begin{equation}
     P(\boldsymbol{x}) = w_o \cdot P_o(\boldsymbol{x}) + w_h \cdot P_h(\boldsymbol{x}),
\end{equation}
where 
\begin{gather}
    w_o + w_h = 1, \nonumber \\
    P_o(\boldsymbol{x}) = softmax(S_c(S_o(\boldsymbol{x}))), \\
    P_h(\boldsymbol{x}) = softmax(S_c(S_h(\boldsymbol{x}))).
    \nonumber
\end{gather}

Here, $w_o$ and $w_h$ represent the respective weights assigned to each branch's prediction. The classification loss is defined as follows:
\begin{equation}
     L_{c} = \mathbb{E}_{(\boldsymbol{x},y) \in \mathcal{D}_p}[\mathcal{L}_{CE}(P(\boldsymbol{x}), y)].
\end{equation}
where $\mathcal{L}_{CE}$ denotes the cross-entropy loss.

Setting $w_o$ and $w_h$ is challenging: if $w_o$ is too large, the original branch may inadvertently learn backdoor knowledge, while a smaller $w_o$ may impede its learning of benign knowledge. Additionally, as indicated in Figure~\ref{fig: motivation}, the shortcut initially captures some benign knowledge. Improper weight settings may cause this portion of benign knowledge to remain permanently embedded in the shortcut, resulting in the original network losing the opportunity to learn it.

To address this issue, we propose a dynamic weight assignment strategy based on prediction entropy. This entropy can be calculated by:
\begin{equation}
     E(\boldsymbol{p}) = -\sum_{k=0}^{K-1} \boldsymbol{p}_k \log \boldsymbol{p}_k  \label{entropy},
\end{equation}
where $\boldsymbol{p}$ is the softmax prediction. For each sample $\boldsymbol{x}$, we first pass it through both branches to obtain predictions $P_o(\boldsymbol{x})$ and $P_h(\boldsymbol{x})$, then assign their contributions to the final prediction as follows:
\begin{equation}
\begin{aligned}
w_o &= \frac{E(P_o(\boldsymbol{x}))}
{E(P_o(\boldsymbol{x})) + E(P_h(\boldsymbol{x}))}, \\
w_h &= \frac{E(P_h(\boldsymbol{x}))}
{E(P_o(\boldsymbol{x})) + E(P_h(\boldsymbol{x}))}.
\end{aligned}
\label{Eq: entropy_based_weight}
\end{equation}

During training, poisoned samples are initially learned by the shortcut branch, inevitably along with a few benign samples. For these poisoned and benign samples, the entropy $E(P_h(\boldsymbol{x}))$ is low, and the weight $w_o$ is high, which gives the original branch an opportunity to learn the benign sample among them. Simultaneously, this weight assignment also enables the original branch to learn from the poisoned ones. To enhance the backdoor knowledge captured by the shortcut branch and prevent it from leaking to the original branch, we design a knowledge decoupling strategy in Section~\ref{knowledge decoupleing}.


\subsection{Knowledge Decoupling}
\label{knowledge decoupleing}
Motivated by the distinct learning behaviors in Figure~\ref{fig: two-subfigures}, we decouple the shortcut and original branches by enforcing constraints in both feature space and final predictions. Through this decoupling, we aim for the original branch to correctly predict only benign samples, while the malicious predictions for poisoned samples are provided by the shortcut branch.

For each sample $\boldsymbol{x}$, we obtain feature vectors $S_o(\boldsymbol{x})$ and $S_h(\boldsymbol{x})$ from the original backbone and honeypot shortcut. We then encourage feature-level decoupling by maximizing their cosine distance:

\begin{equation}
     L_{dpf} = \mathbb{E}_{(\boldsymbol{x},y) \in \mathcal{D}_p} \langle S_o(\boldsymbol{x}), S_h(\boldsymbol{x}) \rangle
     \label{feature decouple}
\end{equation}
where $\langle \cdot,\cdot \rangle$ denotes cosine similarity.
However, feature decoupling alone is insufficient, as the classifier $S_c$ may still adjust itself to accommodate backdoor requirements. Therefore, we introduce an additional prediction-level decoupling loss that jointly constrains the classifier and feature extractors:
\begin{equation}
     L_{dpp} = -\mathbb{E}_{(\boldsymbol{x},y) \in \mathcal{D}_p} {\lVert P_o(\boldsymbol{x}), P_h(\boldsymbol{x}) \rVert}_2
     \label{prediction decouple}
\end{equation}
The cooperation between $L_{dpf}$ and $L_{dpp}$ effectively enables the original and shortcut branches to produce different predictions for each sample. 

However, both losses only consider the decoupling state at the current epoch, and their constraints are satisfied as long as the two branches learn the same sample as different classes. This implies that if the shortcut branch fails to make the malicious prediction for a poisoned sample, the original branch may classify it as the target class due to the classification constraint, resulting in the learning of backdoor knowledge. This issue is more likely when trigger patterns are complex, making it difficult for the shortcut branch to capture stable backdoor knowledge in early epochs. In such cases, the original branch may continuously compete with the shortcut for backdoor knowledge, even snatching it from the shortcut by leveraging its greater learning capacity. 
To address this issue, we leverage previously learned knowledge to guide the current learning direction. Specifically, we construct a feature prototype set for each branch to summarize their knowledge from the previous epoch, as storing full features for all samples is memory-intensive. The prototypes are defined as follows:
\begin{equation}
     \boldsymbol{c}_o^k = \frac{1}{N_k} \displaystyle\sum_{i=1}^{N_k} S_o^{t-1}(\boldsymbol{x}_i^k), \quad 
     \boldsymbol{c}_h^k = \frac{1}{N_k} \displaystyle\sum_{i=1}^{N_k} S_h^{t-1}(\boldsymbol{x}_i^k),
\end{equation}
where $\{\boldsymbol{x}_i^k\}_{i=1}^{N_k}$ denotes samples with label $k$, and $S_o^{t-1}$ and $S_h^{t-1}$ are the original and shortcut branches from the previous epoch. The prototype sets for the two branches are denoted as $\{\boldsymbol{c}_o^k\}_{k=0}^{K-1}$ and $\{\boldsymbol{c}_h^k\}_{k=0}^{K-1}$. We push the two branches to learn in different directions by adding a loss that maximizes the distance between their features and the corresponding prototypes of the other branch:
\begin{equation}
     L_{g} = \mathbb{E}_{(\boldsymbol{x},y) \in \mathcal{D}_p} [ \langle S_o(\boldsymbol{x}), \boldsymbol{c}_h^y \rangle + \langle S_h(\boldsymbol{x}), \boldsymbol{c}_o^y \rangle].
     \label{learning direction guidance}
\end{equation}
Additionally, we enhance the backdoor knowledge in the shortcut branch by encouraging it to learn along its previous direction, which can be interpreted as the backdoor direction:
\begin{equation}
     L_{h} = -\mathbb{E}_{(\boldsymbol{x},y) \in \mathcal{D}_p} \langle S_h(\boldsymbol{x}), \boldsymbol{c}_h^y \rangle.
     \label{shortcut simmilarity}
\end{equation}
Intuitively, $L_h$ functions as an inertial constraint that encourages the shortcut branch to remain on its discovered backdoor learning trajectory, while its combination with $L_g$ imposes an exclusivity constraint that prevents the original branch from “stealing” backdoor knowledge, without hindering its capacity to learn benign representations.

In summary, the training loss can be expressed as follows:
\begin{equation}
     L = L_c + L_g + L_h + L_{dpf} + \alpha \cdot L_{dpp},
     \label{training loss}
\end{equation}
where $\alpha$ controls the strength of decoupling at the prediction level. Specifically, $L_c$ is applied alone during the first epoch to warm up the network, allowing each branch to initially learn its respective knowledge. In the subsequent decoupling, the backdoor knowledge will be channeled into the shortcut branch, while the benign knowledge will flow to the original branch. This behavior is analyzed in Section~\ref{Learning Behaviors Analysis}, and the dynamic of $w_o$ and $w_h$ are further examined in Section~\ref{ablation_study: weight assignment}.
\section{Experiments}
\label{sec:experiment}

\begin{table*}[t]
\belowrulesep=0pt
\aboverulesep=0pt
\centering

\Huge
\renewcommand\arraystretch{1.15}
\resizebox{\linewidth}{!}{
\begin{tabular}{c|c|cc|cc|cc|cc|cc|cc|cc|cc|cc|cc|cc|cc}
    \toprule
    \multirow{2}[4]{*}{Dataset} & \multirow{2}[4]{*}{Attack} & \multicolumn{2}{c|}{No Defense} & \multicolumn{2}{c|}{FP} & \multicolumn{2}{c|}{NAD} & \multicolumn{2}{c|}{ABL} & \multicolumn{2}{c|}{DBD} & \multicolumn{2}{c|}{CBD} & \multicolumn{2}{c|}{ASD} & \multicolumn{2}{c|}{V\&B} & \multicolumn{2}{c|}{PIPD} & \multicolumn{2}{c|}{PDB} & \multicolumn{2}{c|}{ESTI} & \multicolumn{2}{c}{TR (Ours)} \\
\cmidrule{3-26}          &       & BA    & ASR   & BA    & ASR   & BA    & ASR   & BA    & ASR   & BA    & ASR   & BA    & ASR   & BA    & ASR   & BA    & ASR   & BA    & ASR   & BA    & ASR   & BA    & ASR   & BA    & ASR \\
    \midrule
    \multicolumn{1}{c|}{\multirow{12}[8]{*}{CIFAR-10}} & None  & 91.19  & -     & 86.44  & -     & \underline{87.34}  & -     & 70.35  & -     & 70.10  & -     & 84.51  & -     & 85.96  & -     & 84.37      & -     & 85.92  & -     & 82.60  & -     & 82.22  & -     & \textbf{90.37 } & - \\
\cmidrule{2-26}          & BadNet & 90.46  & 99.93  & 87.21  & 5.63  & 86.57  & 1.93  & 86.11  & 3.04  & 86.94  & 1.76  & 87.46  & 1.06  & 84.58  & 1.51  & 88.28  & 1.84  & \underline{89.13 } & 0.64  & 81.60  & 0.21  & 88.92  & \textbf{0.00 } & \textbf{89.85 } & \textbf{0.00 } \\
          & Blended & 90.30  & 98.63  & 86.92  & 6.99  & 75.97  & 5.48  & 85.34  & 16.23  & 86.83  & 5.12  & 87.48  & 1.96  & 86.51  & 2.02  & \underline{88.15 } & 2.08  & 87.35  & 2.69  & 80.90  & \underline{0.95 } & 87.54  & 64.54  & \textbf{89.06 } & \textbf{0.01 } \\
          & WaNet & 89.71  & 92.27  & 85.87  & 2.62  & \underline{86.97 } & 4.21  & 75.74  & 22.24  & 84.60  & 5.86  & 86.55  & 4.24  & 84.85  & 2.03  & 84.21  & 3.34  & 86.39  & 4.73  & 81.16  & \underline{1.80 } & 83.55  & 12.02  & \textbf{88.90 } & \textbf{1.33 } \\
          & Dynamic & 89.78  & 85.03  & 86.59  & 10.49  & 87.07  & 2.70  & 85.34  & 18.46  & 85.42  & 10.21  & 85.67  & 0.86  & 83.64  & 4.49  & 81.92  & 47.63  & 89.52  & \underline{0.63 } & 80.40  & 1.92  & \underline{89.60 } & \textbf{0.23 } & \textbf{90.01 } & 1.06  \\
          & DataFree & 90.82  & 99.96  & 86.53  & 2.07  & 78.38  & 3.16  & 80.13  & 0.96  & 71.16  & 13.41  & 85.75  & 2.00  & 85.95  & 2.89  & 88.17  & 1.41  & 85.27  & 0.25  & 81.37  & 0.53  & \textbf{90.50 } & \textbf{0.00 } & \underline{90.27 } & \textbf{0.00 } \\
          & DUBA  & 89.66  & 95.31  & 80.77  & 26.89  & 86.46  & 43.85  & 65.55  & 99.56  & 69.78  & 12.73  & 87.13  & 2.27  & 86.04  & 2.88  & 84.55  & 4.18  & \underline{87.38 } & 5.19  & 80.71  & 2.18  & 20.01  & \textbf{0.01 } & \textbf{88.18 } & \underline{1.58 } \\
          & SIG   & 90.85  & 70.50  & 86.72  & 13.62  & 80.44  & 0.63  & 73.19  & 0.33  & 69.52  & 13.38  & 86.60  & 26.03  & 86.29  & 3.07  & 85.53  & 3.46  & 87.83  & 0.70  & 82.12  & \underline{0.07 } & \textbf{90.97 } & 0.13  & \underline{89.42 } & \textbf{0.00 } \\
          & CL    & 85.34  & 95.44  & 83.33  & 19.17  & 84.89  & 8.89  & 86.18  & 1.86  & 69.97  & 3.61  & 87.46  & 1.17  & 84.65  & 3.93  & 85.37  & 2.84  & 85.12  & 2.39  & 82.51  & 1.76  & \underline{88.74 } & \textbf{0.26 } & \textbf{89.33 } & \underline{0.74 } \\
          & Narcissus & 90.73  & 95.53  & 85.47  & 69.03  & 87.34  & 44.20  & 71.95  & 45.66  & 70.02  & 99.74  & 84.98  & 26.43  & 82.49  & 98.58  & 84.69  & 32.41  & \underline{89.55 } & 6.81  & 82.37  & 1.25  & 89.15  & \underline{0.85 } & \textbf{89.90 } & \textbf{0.53 } \\
          & PCBA  & 90.11  & 99.51  & 86.25  & 51.59  & 84.88  & 26.46  & 65.74  & 97.74  & 68.12  & 98.61  & 85.54  & 81.84  & 81.76  & 99.97  & 66.93  & \textbf{0.00 } & 80.92  & 10.38  & 82.01  & 0.94  & \textbf{89.73 } & 1.44  & \underline{89.05 } & \underline{0.41 } \\
\cmidrule{2-26}          & \textbf{Average} & 89.78  & 93.21  & 85.57  & 20.81  & 83.90  & 14.15  & 77.53  & 30.61  & 76.24  & 26.44  & 86.46  & 14.79  & 84.68  & 22.14  & 83.78  & 9.92  & \underline{86.85}  & 3.44  & 81.52  & \underline{1.16}  & 81.87  & 7.95  & \textbf{89.40 } & \textbf{0.57 } \\
    \midrule
    \multicolumn{1}{c|}{\multirow{9}[8]{*}{GTSRB}} & None  & 99.33  & -     & 94.94  & -     & 97.01  & -     & 83.90  & -     & 91.16  & -     & 75.45  & -     & 96.12  & -     & \underline{98.05 } & -     & 93.88  & -     & 92.91  & -     & 97.87      & -     & \textbf{98.90 } & - \\
\cmidrule{2-26}          & BadNet & 98.64  & 100.00  & 96.87  & 0.03  & \underline{96.94 } & 0.06  & 92.58  & 0.03  & 86.13  & \textbf{0.00 } & 84.65  & 2.32  & 96.52  & 0.10  & 88.72  & 0.35  & 96.50  & 0.01  & 92.77  & 0.02  & 96.72  & \textbf{0.00 } & \textbf{98.57 } & \textbf{0.00 } \\
          & Blended & 98.62  & 99.84  & 95.03  & 12.88  & 93.46  & 1.30  & 89.07  & 5.52  & 86.25  & 99.98  & 76.05  & 88.18  & \underline{96.21 } & \underline{0.12 } & 75.84  & 48.38  & 94.59  & 0.83  & 92.61  & 0.29  & 95.02  & \textbf{0.06 } & \textbf{97.88 } & 0.62  \\
          & WaNet & 96.18  & 93.53  & \underline{96.52 } & 18.08  & 96.17  & 67.20  & 94.50  & 38.29  & 84.71  & \textbf{0.05 } & 90.50  & 86.11  & 95.91  & 42.26  & 83.16  & 7.50  & 95.04  & 3.48  & 89.42  & 0.77  & 93.84  & 0.84  & \textbf{96.59 } & \textbf{0.05 } \\
          & DataFree & 97.04  & 100.00  & 96.90  & 0.03  & 97.37  & 0.03  & 95.21  & 0.03  & 92.71  & 0.38  & 76.18  & 2.56  & 95.05  & 0.09  & \underline{98.31 } & 0.14  & 96.42  & 0.22  & 92.20  & 0.18  & 97.22  & \textbf{0.00 } & \textbf{98.80 } & \textbf{0.00 } \\
          & DUBA  & 98.83  & 93.91  & 95.48  & 20.81  & 93.26  & 8.61  & 94.04  & 3.84  & 88.09  & 0.05  & 90.86  & 92.85  & 95.95  & \textbf{0.03 } & 90.77  & 9.65  & \underline{96.35 } & 4.37  & 90.62  & 3.57  & 80.21  & 3.70  & \textbf{98.58 } & \textbf{0.03 } \\
          & SIG   & 98.94  & 100.00  & 94.58  & 14.19  & 88.78  & \underline{0.02 } & 86.90  & 99.88  & 86.84  & 69.82  & 89.84  & 24.17  & 96.26  & 45.93  & 91.28  & 2.80  & 92.39  & 5.27  & 93.16  & 59.75  & \textbf{97.49 } & 9.04  & \underline{97.17 } & \textbf{0.00 } \\
          & Narcissus & 98.50  & 67.40  & 94.61  & 18.56  & \underline{97.67 } & 5.83  & 87.46  & 32.83  & 92.58  & 0.45  & 90.87  & 58.37  & 94.60  & 93.75  & 97.64  & 36.71  & 93.74  & 3.61  & 92.82  & 13.80  & 97.56  & \textbf{0.00 } & \textbf{98.99 } & \textbf{0.00 } \\
\cmidrule{2-26}          & \textbf{Average} & 98.11  & 93.53  & 95.71  & 12.08  & 94.81  & 11.86  & 91.39  & 25.77  & 88.19  & 24.39  & 85.56  & 50.65  & \underline{95.79 } & 26.04  & 89.39  & 15.08  & 95.00  & 2.54  & 91.94  & 11.20  & 94.01  & \underline{1.95 } & \textbf{98.08 } & \textbf{0.10 } \\
    \bottomrule
    \end{tabular}%
}

\caption{The performance of 11 defense methods against multiple backdoor attacks on two datasets. The best results are highlighted in \textbf{bold}, and the second-best results are \underline{underlined}. \emph{None} denotes the totoally clean datasets.}
\label{defense against classical attack}
\end{table*}

\subsection{Experimental Settings}
\paragraph{Datasets and Models.}
In the main paper, we evaluate our defense on CIFAR-10 with WRN-16-1 and on GTSRB~\cite{GTSRB} with ResNet-18 to demonstrate effectiveness across architectures. To further evaluate generalization across datasets and architectures, we include additional experiments on CIFAR-100~\cite{CIFAR}, ImageNet-1k~\cite{deng2009imagenet}, as well as on PreActResNet-18~\cite{preact-resnet}, VGG-16~\cite{VGG}, and ViTs~\cite{ViT} in \emph{Appendix}~\ref{Appendix: Evaluation on Additional Datasets and Model Architectures}. Dataset details are provided in \emph{Appendix}~\ref{Appendix: datasets}.

\paragraph{Attack Configures.}


We consider ten state-of-the-art backdoor attacks, including six dirty-label methods: BadNets, Blend~\cite{blend}, WaNet~\cite{wanet}, Dynamic~\cite{Dynamic}, DataFree~\cite{datafree}, and DUBA~\cite{DUBA}, and four clean-label attacks: SIG~\cite{SIG}, CL~\cite{CL}, Narcissus~\cite{narcissus}, and PCBA~\cite{PCBA}. All attacks are implemented according to their released code to ensure strong performance. Target labels are set to 0 for CIFAR-10 and 1 for GTSRB, with a default poisoning rate of 10\%. Implementation details are provided in \emph{Appendix}~\ref{Appendix: attacks}.

\paragraph{Defense Configures and Training Details.}
We compare our method with ten state-of-the-art defenses: FP~\cite{FP}, NAD~\cite{NAD}, ABL~\cite{ABL}, DBD~\cite{DBD}, CBD~\cite{CBD}, ASD~\cite{ASD}, V\&B~\cite{zhu2023victim}, PIPD~\cite{PIPD}, PDB~\cite{PDB}, and ESTI~\cite{ESTI}. FP and NAD require 5\% local benign samples, while ASD, PDB, and ESTI use 0.2\%, 10\%, and 1\% of the training data as seed samples, respectively. Our model is trained for 200 epochs using SGD with an initial learning rate of 0.1, decayed by a factor of 10 every 50 epochs. The weighting factor $\alpha$ is set to 1 for CIFAR-10 and linearly decreased from 2 to 1 over the first 50 epochs for GTSRB.

\paragraph{Evaluation Metrics.}

We evaluate the defense performance using two metrics: Benign Accuracy (BA) on clean data and Attack Success Rate (ASR) on poisoned data. The lower the ASR and the higher BA, the more effective the defense.

\subsection{Effectiveness and Efficiency}

We compare TR with ten SOTA defenses in Table~\ref{defense against classical attack}. \emph{No Defense} serves as the attack baseline. Overall, TR achieves the best average BA and ASR on both datasets, even compared with defenses requiring benign samples. It consistently reduces ASR to below or near 1\% (often 0\%) while maintaining high BA, especially on clean datasets. TR also remains robust under different poisoning rates, as shown in \emph{Appendix}~\ref{Appendix: poisoning rate}.

Defenses based on data partitioning, such as ABL, DBD, ASD, V\&B, and ESTI, inevitably treat a portion of the training samples as poisoned, even when the dataset is clean, which leads to degraded BA. Moreover, all other defenses can be ineffective under certain attack settings, resulting in higher ASR or reduced BA. As for TR, when the data is clean, the shortcut branch captures only a few of easily learned benign information and does not interfere with the main branch in learning generalizable features from harder samples. When the data is poisoned, the shortcut rapidly absorbs backdoor knowledge, thereby preventing the main network from learning it. We further analyze class-wise BA in \emph{Appendix}~\ref{Appendix: Class-wise Benign Performance}, showing that TR has minimal impact on benign performance. 



We report the training cost of TR in \emph{Appendix} Table~\ref{classical running time}, showing that TR is more efficient than others. This efficiency stems from introducing only a shallow shortcut, without repeated data partitioning or iterative multi-network training.

Results on a broader range of dataset–architecture combinations, reported in \emph{Appendix} Tables~\ref{defense on CIFAR-100}, \ref{defense on ImageNet-1K}, \ref{defense on CIFAR-10 with ResNet-18}, \ref{defense on GTSRB with WRN-16-1}, and \ref{Comparison with NLP-focused Defense on CIFAR-10}, further demonstrates the generalization capability of TR, including our adaptive shortcut generation strategy. Notably, the Table~\ref{defense on ImageNet-1K} and \ref{Comparison with NLP-focused Defense on CIFAR-10} results also verify that our method remains effective against fine-tuning attacks.

\begin{figure}[t]
  \centering 
  \includegraphics[width=\linewidth]{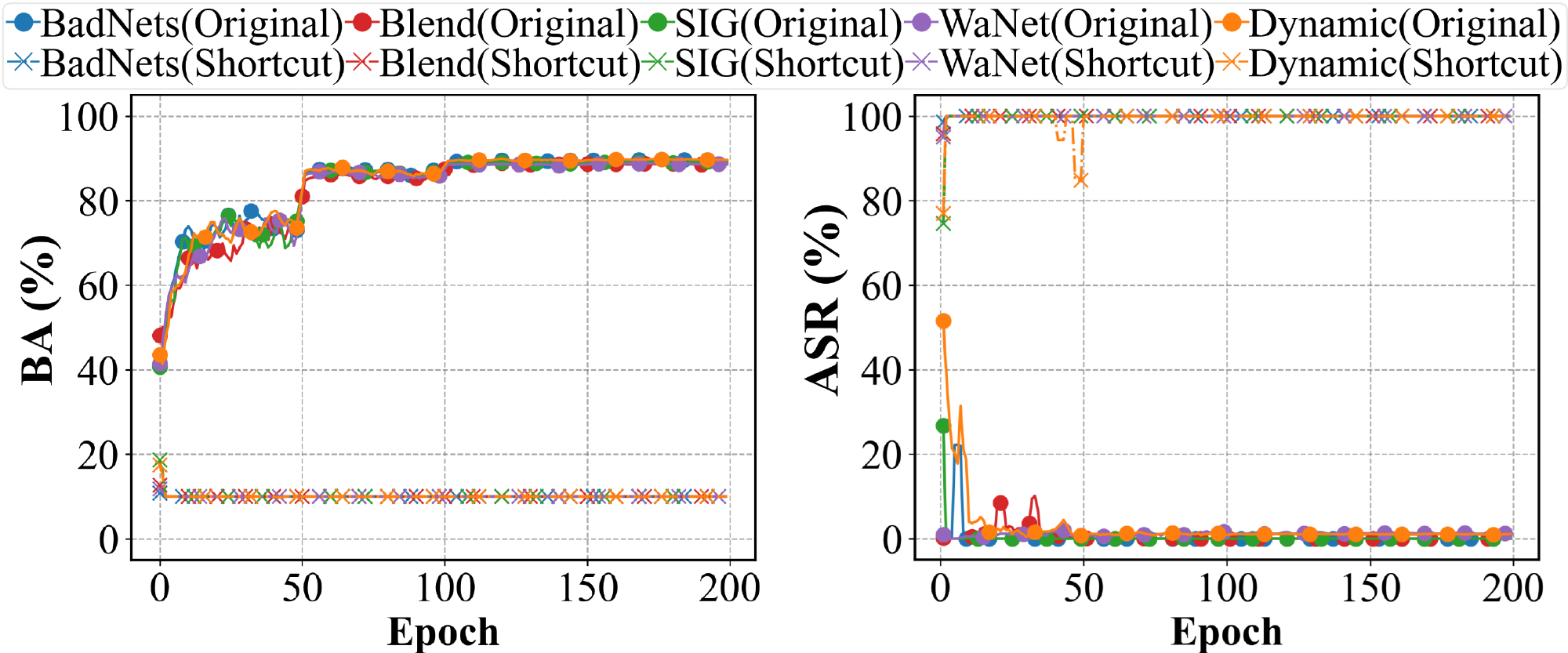}
\caption{The BA and ASR of both branches during the decoupling training process on CIFAR-10 under the same settings in Table~\ref{defense against classical attack}.}
\label{fig: BA&ASR}
\end{figure}

\subsection{Performance of the Two Branches}
\label{Learning Behaviors Analysis}
To better illustrate the decoupling process, Figure~\ref{fig: BA&ASR} presents BA and ASR of the original and shortcut branches. The original branch's BA steadily increases, while that of the shortcut decreases, with both eventually stabilizing. Although the shortcut retains a small amount of benign knowledge (BA $\approx$ 10\%), it does not impede the original branch due to entropy-based weight assignment. In early epochs, the shortcut's ASR is markedly higher than the original’s, indicating it captures most backdoor knowledge. As training progresses, the original branch’s ASR rapidly drops to 0\%, while the shortcut remains at 100\%, demonstrating that the shortcut effectively traps backdoor behavior and shields the original branch.

\begin{table*}[t]
\belowrulesep=0pt
\aboverulesep=0pt
\centering
\resizebox{0.9\linewidth}{!}{
\begin{tabular}{cc|cc|cccccccc|cc|c}
    \toprule
    \multicolumn{4}{c|}{Choices}  & \multicolumn{2}{c}{BadNets} & \multicolumn{2}{c}{Blend} & \multicolumn{2}{c}{WaNet} & \multicolumn{2}{c|}{Dynamic} & \multicolumn{2}{c|}{SIG} & Mean \\
    \midrule
    $L_{dpf}$ & $ L_{dpp}$ & $L_h$ & $L_g$ & BA    & ASR   & BA    & ASR   & BA    & ASR   & BA    & ASR   & BA    & ASR   & ASR \\
    \midrule
          &       &       &       & 90.77\% & 99.91\% & 90.20\% & 98.46\% & 88.59\% & 92.92\% & 89.78\% & 85.86\% & 90.88\% & 76.80\% & 90.79\% \\
    \midrule
          &       & \checkmark     & \checkmark     & {90.93\%} & \st{99.70\%} & 90.36\% & \st{97.96\%} & 88.65\% & \st{93.10\%} & 89.55\% & \st{82.01\%} & {90.41\%} & \st{70.72\%} & 88.70\% \\
    \checkmark     &       & \checkmark     & \checkmark     & 90.56\% & \st{96.84\%} & {90.37\%} & \st{97.72\%} & {88.97\%} & \st{92.04\%} & 89.55\% & \st{83.24\%} & 90.14\% & \st{75.82\%} & 89.13\% \\
          & \checkmark     & \checkmark     & \checkmark     & 89.74\% & {0.00\%} & 89.10\% & {0.00\%} & 88.49\% & \st{82.81\%} & 89.15\% & \st{65.13\%} & 89.65\% & \st{73.37\%} & 44.26\% \\
    \midrule
    \checkmark     & \checkmark     &       &       & 90.16\% & \st{96.98\%} & 89.26\% & {0.00\%} & 88.64\% & \st{83.59\%} & 89.05\% & \st{88.34\%} & 89.65\% & {0.00\%} & 36.30\% \\
    \checkmark     & \checkmark     & \checkmark     &       & 89.53\% & \st{99.69\%} & 88.38\% & {0.00\%} & 88.15\% & \st{83.64\%} & 88.97\% & \st{66.72\%} & 89.60\% & {0.00\%} & 30.07\% \\
    \checkmark     & \checkmark     &       & \checkmark     & 89.93\% & {0.00\%} & 88.77\% & {0.00\%} & 88.48\% & \st{83.84\%} & 89.06\% & \st{83.67\%} & 89.93\% & \st{80.94\%} & 49.69\% \\
    \midrule
    \checkmark     & \checkmark     & \checkmark     & \checkmark     & 89.83\% & {0.00\%} & 89.06\% & 0.01\% & 88.90\% & {1.33\%} & {90.01\%} & 1.06\% & 89.42\% & {0.00\%} & {0.48\%} \\
    \bottomrule
    \end{tabular}%
}
\caption{Ablation study of the four losses on CIFAR-10. The deleted cases indicate \st{defense failures}. It is evident that combining all four losses is essential for achieving robust defense performance.}
\label{ablation study}
\end{table*}

\subsection{Ablation Studies}
\label{ablation_study}

\paragraph{Weight Assignment.}
\label{ablation_study: weight assignment}
To evaluate our entropy-based weight assignment, we replace it with several fixed schemes in Table~\ref{fixed weight assignment}. Assigning equal weights (0.5) to both branches reduces ASR to 0\%, but results in lower benign accuracy, as benign knowledge captured by the shortcut is not learned by the original branch. To address this, we gradually increase $w_o$ and decrease $w_h$, which improves BA (\eg\ with 0.8/0.2 and 0.9/0.1) but also raises ASR. This highlights the difficulty of balancing BA and ASR using fixed weights. Without the entropy-based weight assignment, certain benign knowledge may be sacrificed and permanently retained in the shortcut branch to obtain a clean original network. In the last column, we reverse the roles of $w_o$ and $w_h$ in Equation~\ref{Eq: entropy_based_weight}, which also lowers BA. As indicated in Figure~\ref{fig: motivation}, the shortcut inevitably captures some easy benign samples early on. In the reversed setting, a consistently higher $w_h$ limits the original branch from learning these samples, resulting in BA degradation.

\begin{figure}[t]
\centering 
\includegraphics[width=0.95\linewidth]{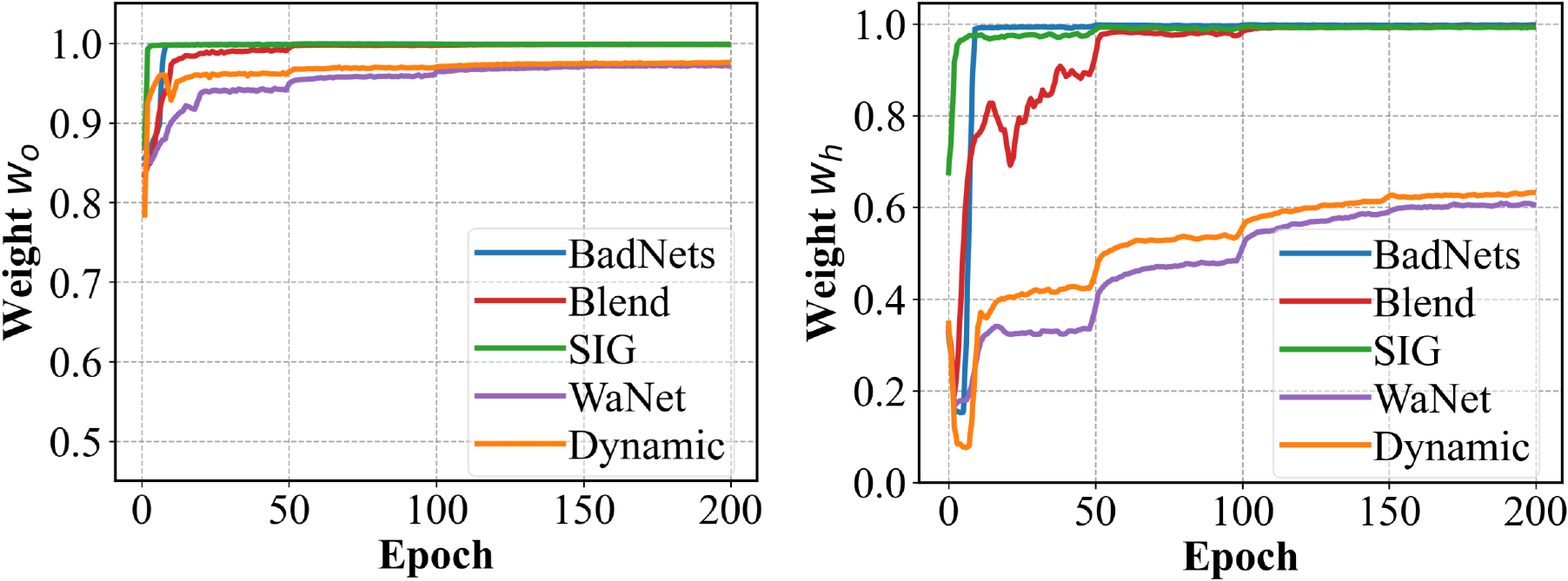}
\caption{Visualization of $w_o$ for benign samples and $w_h$ for poisoned samples during decoupling training on CIFAR-10.}
\label{fig: weight analysis}
\end{figure}

Figure~\ref{fig: weight analysis} illustrates the weight dynamics in our defense. For benign samples, $w_o$ starts high and quickly reaches 1, as the shortcut initially learns little benign knowledge constrained by its limited capacity. The decoupling loss then drives the shortcut toward confidently incorrect predictions, reflected in its low BA in Figure~\ref{fig: BA&ASR}.

For poisoned samples, $w_h$ is initially low because the shortcut strongly learns the backdoor, confidently predicting poisoned samples as the target label, as reflected in its high initial ASR in Figure~\ref{fig: BA&ASR}. Given the low $w_h$, poisoned samples are learned primarily through the original branch. However, the decoupling loss $L_{dpp}$ prevents the original branch from mimicking the shortcut’s malicious predictions, increasing the uncertainty and entropy of the original branch for poisoned samples, which further reduces $w_h$. Since the shortcut possesses strong backdoor knowledge and is reinforced by $L_h$, it is easier to shift the predictions of the original branch for poisoned samples than those of the shortcut. Consequently, the original branch increasingly predicts non-target labels for poisoned samples with high confidence (evidenced by its reduced ASR in Figure~\ref{fig: BA&ASR}), as detailed in the case study in \emph{Appendix}~\ref{Appendix: Case Study}.


To better clarify these weight dynamics, we provide a detailed analysis in \emph{Appendix}~\ref{Appendix: weight assignment}, and visualize the decoupling process in \emph{Appendix}~\ref{Appendix: learning behavior}.

\paragraph{Loss Effect.}
In Table~\ref{ablation study}, we examine the impact of each additional loss on the defense performance. First, we train the network without any additional losses as the baseline in row 3, which fails to defend against all attacks. Rows 4$\sim$6 reveal that the defense is ineffective in most cases when either \( L_{dpf} \) or \( L_{dpp} \) is omitted, highlighting the necessity of decoupling both in feature space and prediction results. Relaying solely on feature decoupling fails in all cases, as the classifier can adapt to satisfy the backdoor requirements. Similarly, decoupling only predictions proves ineffective against complex attacks, such as WaNet and Dynamic (noisy mode). Without constraints in the feature space, the original branch eventually learns these complex patterns, even if the shortcut captures them initially, thereby undermining defense performance.

Rows 7$\sim$9 explore the role of learning direction guidance losses $L_h$ and $L_g$, where all cases against WaNet and Dynamic fail. For these two complex attacks, $L_h$ allows the shortcut sufficient epochs to capture stable backdoor knowledge, while $L_g$ helps prevent the original branch from learning leaked backdoor knowledge during this period by preventing it from learning along the poisoned direction. Without $L_g$, although the shortcut initially learns backdoor knowledge well against BadNets, the original branch may still become contaminated by several leaked poisoned samples. For the clean-label attack SIG, backdoor knowledge is more delicate due to competition between the trigger and the target-class ground-truth pattern. Here, strengthening the shortcut's captured knowledge is critical, as applying $L_g$ alone will disrupt the backdoor knowledge in the shortcut, causing too many poisoned samples to leak to the original branch. Ultimately, all four losses are required for a robust defense. The selection of loss weight $\alpha$ is provided in \emph{Appendix}~\ref{Appendix: weight alpha}.

\begin{table}[t]
\belowrulesep=0pt
\aboverulesep=0pt
\centering
\resizebox{\linewidth}{!}{
\begin{tabular}{c|cccccc}
    \toprule
    $w_o / w_h$  & 0.5/0.5 & 0.6/0.4 & 0.7/0.3 & 0.8/0.2 & 0.9/0.1 & $w_h / w_o$\\
    \midrule
    BA    & 82.22\% & 81.90\% & 82.04\% & 85.09\% & 88.71\% & 88.78\%\\
    ASR   & 0.00\% & 0.00\% & 0.00\% & 47.70\% & 92.28\% & 0.00\%\\
    \bottomrule
    \end{tabular}
}
\caption{Performance under different weight assignments on the BadNets attack. The case `0.9/0.1' indicates $w_o=0.9$ and $w_h=0.1$. The last case indicates switching the assignment for $w_o$ and $w_h$.}
\label{fixed weight assignment}
\end{table}

\section{Conclusions}
\label{sec:conclusion}

We reveal that backdoor knowledge can be naturally captured by a parallel shortcut branch and propose TR, a simple and effective defense. By introducing a honeypot shortcut and a decoupling mechanism, TR isolates backdoor knowledge from benign ones, enabling backdoor removal via post-training shortcut pruning without requiring additional data. We further incorporate an adaptive shortcut generation strategy to enhance generalization. Extensive experiments demonstrate the effectiveness and robustness of our TR. Future work will extend TR to counter optimized attacks with higher semantic complexity that may challenge the shortcut-based capture.


\appendix
\clearpage





\bibliographystyle{named}
\bibliography{ijcai26}

\clearpage



\section{Detailed Settings}
\label{Appendix: Detailed Experimental Settings}

All experiments were conducted on a workstation equipped with an RTX 3090 GPU (24GB) and 64GB of RAM.

\subsection{Details about Datasets}
\label{Appendix: datasets}
The detailed dataset information is summarized in Table~\ref{Table: datasets}.

\begin{table}[htbp]
\belowrulesep=0pt
\aboverulesep=0pt
\centering
\Huge
\renewcommand\arraystretch{1.1}
\begin{center}
\resizebox{\linewidth}{!}{
    \begin{tabular}{c|c|c|c|c}
    \toprule[3pt]
    Dataset & Classes & Input Size  &Training Images &Test Images \\
    \hline 
    CIFAR-10 &10 & 32 x 32 x 3 & 50000 & 10000  \\
    \hline 
    GTSRB &{43} & {32 x 32 x 3} & {39209} & {12630} \\
    \hline 
    CIFAR-100 &{100} & {32 x 32 x 3} & {50000} & {10000} \\
    \hline
    ImageNet-1k &{1000} & {224 x 224 x 3} & {1281167} & {50000} \\
    
    \bottomrule[3pt]
    \end{tabular}
}
\end{center}
\caption{Detailed information of the datasets used in our experiments.}
\label{Table: datasets}
\end{table}

\subsection{Details about Attacks}
\label{Appendix: attacks}
We trained all attack baselines for 200 epochs using the SGD optimizer with an initial learning rate of 0.1, a weight decay of \(1 \times 10^{-4}\), and a momentum of 0.9. The learning rate is reduced by a factor of 10 every 50 epochs. We use a batch size of 128 for CIFAR-10, GTSRB, and CIFAR-100, and a batch size of 64 for ImageNet-1k. The poisoned samples are randomly selected.


\subsection{Settings for Learning Behaviors Analysis}
\label{Appendix: Detailed Settings for Learning Behaviors Analysis}
The introduced shortcut consists of two \emph{Conv-Attention-BN-ReLU} blocks, as described in Section~\ref{sec: Honeypot Shortcut Designing}. The final output of the modified WRN-16-1 network is computed as:
\begin{equation}
P(\boldsymbol{x}) = 0.5 \cdot P_o(\boldsymbol{x}) + 0.5 \cdot P_h(\boldsymbol{x}),
\end{equation}
We train the modified network for 5 epochs using stochastic gradient descent (SGD) with a momentum of 0.9, weight decay of 0.0005, and an initial learning rate of 0.1. The batch size is set to 128.

In Figure~\ref{fig: feature_vis_two_branches}, we visualize 2,500 randomly selected training samples using t-SNE.

\begin{table}[t]
\belowrulesep=0pt
\aboverulesep=0pt
\centering

\resizebox{\linewidth}{!}{
    \begin{tabular}{cccccccc}
    \toprule
    Normal & DBD   & CDB & ASD   & V\&B & PDB & ESTI & TR(Ours) \\
    \midrule
    1052 & 13691 & 2159& 8858 & 7750 & 3360 & 5309 & 1332 \\
    \bottomrule
    \end{tabular}%
}
\caption{Time (seconds) of four defenses versus normal training, evaluated on CIFAR-10 with WRN-16-1.} 
\label{classical running time}
\end{table}

\subsection{Adaptive Shortcut Designing}
\label{Appendix: Adaptive Shortcut Designing}
For CNNs, the shortcut branch is composed of repeated \emph{Conv-Attention-BN-ReLU} blocks. Each block begins with a $3\times3$ convolution with stride 2 to perform $2\times$ downsampling. The attention layer is implemented as a 1-channel convolution that maintains the spatial dimensions of the input and produces an attention map $M$. This map is applied element-wise to the convolution output $f$, as described in Equation~\ref{attention}. The number of such blocks is determined by the number of downsampling stages in the original network. For example, the WRN-16-1 backbone performs two 2× downsamplings, so we construct the shortcut branch using two \emph{Conv-Attention-BN-ReLU} blocks to align its spatial resolution with that of the backbone. To facilitate effective decoupling, we also match the output channel dimensions between the shortcut and backbone branches.

For ViTs, we construct the shortcut branch as a lightweight transformer network consisting of half the number of blocks as in the original architecture. The performance is not highly sensitive to this choice—other ratios such as 1/4 or 3/4 also achieve competitive results, as shown in Table~\ref{Ablation study of the shortcut size for ViT}.

\section{Gradient-Based Analysis and Verification of Learning Behaviors}
\label{Appendix: perspective of gradient}

\subsection{Theoretical Analysis}
To understand why poisoned samples are preferentially captured by the honeypot shortcut, we analyze the learning dynamics through the lens of gradient flow. Let $\boldsymbol{x}$ be an input sample with label $y$. We denote the original (deep) backbone and the shortcut (shallow) branch as $S_o(\cdot; \theta_o)$ and $S_h(\cdot; \theta_h)$, producing feature vectors $\boldsymbol{f}_o \in \mathbb{R}^d$ and $\boldsymbol{f}_h \in \mathbb{R}^d$, respectively. A shared classifier $S_c(\cdot; \phi)$, parameterized by weights $W$ and bias $b$, computes the logits:

\begin{equation} 
\boldsymbol{z}_o = W \boldsymbol{f}_o + b, \quad \boldsymbol{z}_h = W \boldsymbol{f}_h + b. 
\end{equation}

The final prediction is the weighted average of the softmax outputs from both branches:
\begin{equation}
P(\boldsymbol{x}) = w_o \cdot \mathrm{softmax}(\boldsymbol{z}_o) + w_h \cdot \mathrm{softmax}(\boldsymbol{z}_h),
\end{equation}
where $w_o + w_h = 1$ (initialized as $0.5$). We minimize the standard cross-entropy loss $L = -\log P_y(\boldsymbol{x})$. Applying the chain rule, the parameter updates for each branch are governed by:

\begin{equation} 
\nabla_{\theta_o} L = w_o \cdot \nabla_{\boldsymbol{z}_o} L \cdot W \cdot \frac{\partial \boldsymbol{f}_o}{\partial \theta_o}, 
\nabla_{\theta_h} L = w_h \cdot \nabla_{\boldsymbol{z}_h} L \cdot W \cdot \frac{\partial \boldsymbol{f}_h}{\partial \theta_h}. \label{eq:gradient_flow} 
\end{equation}

The shortcut's learning preference is driven by a cascade of three factors: the consistency of optimization signals provides the basis, structural simplicity (Simplicity Bias) enables rapid capture, and attention-induced gradient focusing significantly amplifies this dominance.


\begin{figure*}[tbp]
\begin{center}
\includegraphics[width=0.99\linewidth]{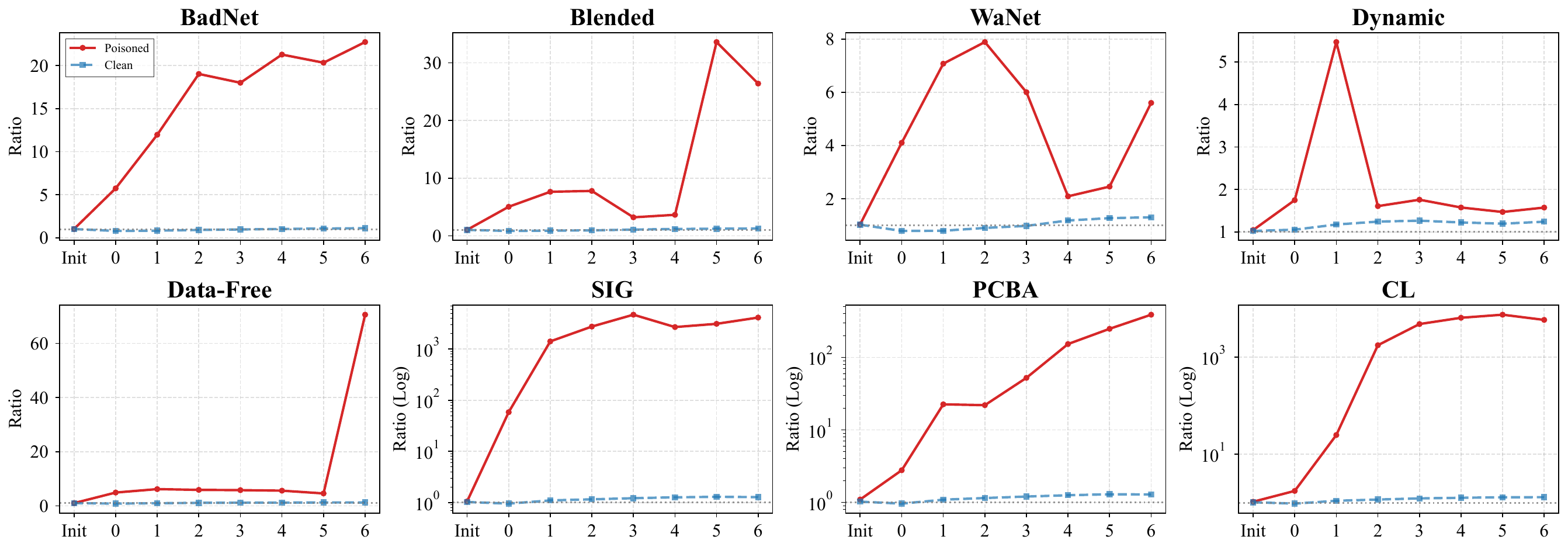}
\end{center}
\vspace{-0.3cm}
   \caption{\textbf{Consistent gradient dominance dynamics during early training stages.}
We track the evolution of the \textbf{logit gradient norm ratio} ($\|\nabla_{\boldsymbol{z}_h} L\| / \|\nabla_{\boldsymbol{z}_o} L\|$) from Initialization (Init) to Epoch 6 across 8 attacks.
\textbf{Divergence:} Despite starting from a balanced state (Init $\approx 1.0$), the shortcut branch rapidly establishes a dominant gradient flow for poisoned samples (Red line), while the ratio for clean samples (Blue line) remains constant at $\approx 1.0$. This confirms the specific structural sensitivity of the shortcut to trigger patterns.
\textbf{Magnitude:} Note that for clean-label attacks like SIG, PCBA, and CL (bottom row), the y-axis is presented in \textbf{log scale} to visualize the exponential surge in gradient magnitude (reaching $>10^3$). This massive disparity indicates a ``winner-take-all'' mechanism where the optimization signal is almost exclusively channeled through the shortcut branch.}
\label{fig:grad_dynamics}
\end{figure*}

\subsubsection{Consistent Optimization Signal via Trigger Patterns}
\label{sec:consistency}

Prior work~\cite{ABL,zhu2023victim} suggests that poisoned samples converge faster due to the high consistency of trigger patterns. In our dual-branch architecture, this consistency creates a stable optimization signal that propagates to both branches. 
Since the classifier head is randomly initialized and shared with balanced weights (0.5), the error gradients with respect to the logits, $\nabla_{\boldsymbol{z}_o} L$ and $\nabla_{\boldsymbol{z}_h} L$, are highly synchronized and comparable in magnitude (as empirically verified in Sec.~\ref{sec:verification}). 
However, for poisoned samples, the invariant trigger pattern causes these gradients to \textbf{accumulate constructively} across the training batch, which can be approximated as:
\begin{equation}
\sum_{\boldsymbol{x} \in \mathcal{B}_{poison}} \nabla_{\boldsymbol{z}} L(\boldsymbol{x}) \approx N \cdot \boldsymbol{g}_{trigger},
\end{equation}
whereas gradients for diverse benign features exhibit higher variance and partial cancellation. This means that both branches effectively receive a low-variance, high-strength optimization signal from the trigger. The critical divergence in learning outcomes, therefore, does not stem from a disparity in these error signals initially, but from how each branch's structure responds to this consistent push (as detailed below).

\subsubsection{Attention-Induced Gradient Focusing}

The magnitude of the update in Eq.~\eqref{eq:gradient_flow} depends on the Jacobian terms $\frac{\partial \boldsymbol{f}}{\partial \theta}$. Our shortcut branch $S_h$ explicitly incorporates an attention mechanism to enhance sensitivity to trigger patterns. As defined in our method, the feature refinement is given by $\boldsymbol{f}^{\prime}_h = \mathrm{ReLU}(\boldsymbol{M}) \otimes \boldsymbol{f}_h$, where $\boldsymbol{M}$ is the attention map. The introduction of this multiplicative interaction creates a gradient focusing effect. During backpropagation, the gradient is modulated by the attention map:

\begin{equation}
\frac{\partial L}{\partial \boldsymbol{f}_h} = \frac{\partial L}{\partial \boldsymbol{f}^{\prime}_h} \otimes \underbrace{\mathrm{ReLU}(\boldsymbol{M})}_{\text{Gating Term}} + \dots
\end{equation}

For poisoned samples, the backdoor trigger represents a consistent and discriminative pattern compared to the varying semantic background. The attention mechanism rapidly learns to activate on these salient regions (i.e., $\boldsymbol{M}_{trigger} > 0$) while suppressing background noise ($\boldsymbol{M}_{bg} \approx 0$). Thus, the term $\mathrm{ReLU}(\boldsymbol{M})$ acts as a gate that amplifies the gradient flow specifically for trigger patterns while blocking gradients from irrelevant regions.

\begin{figure}[tbp]
\begin{center}
\includegraphics[width=0.99\linewidth]{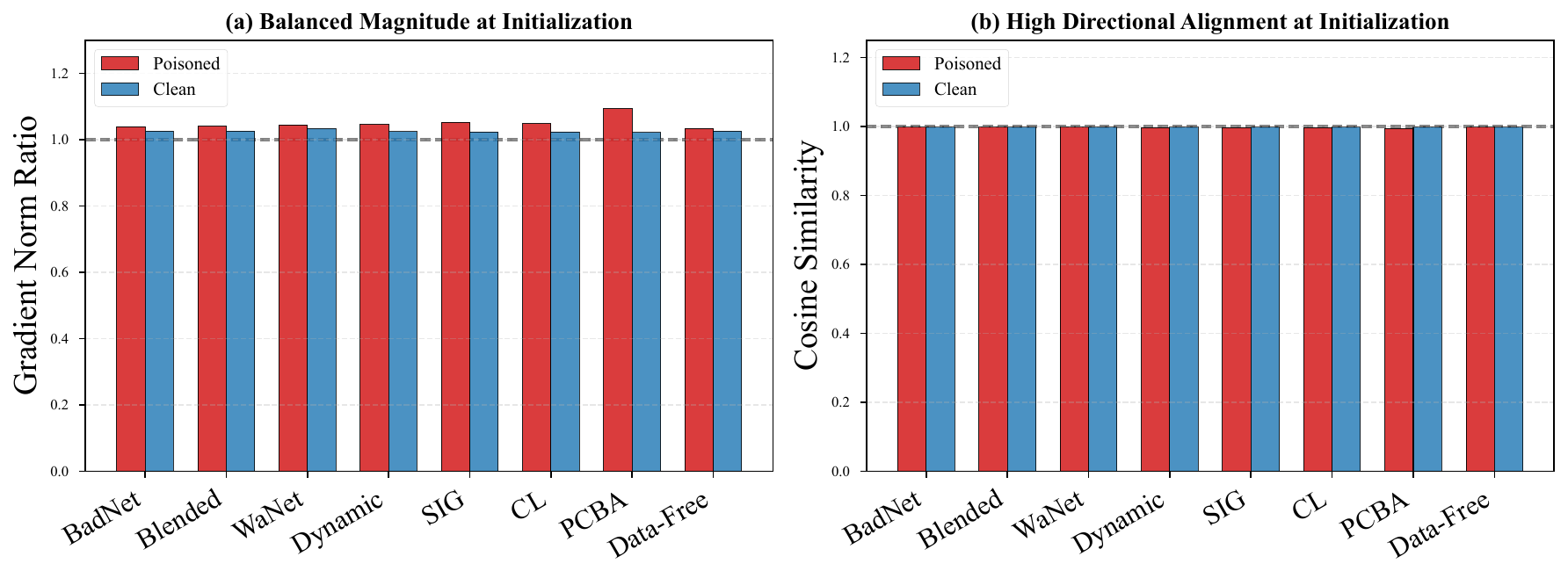}
\end{center}
   \caption{\textbf{Empirical verification of balanced initialization across 8 attacks.}
We measured the gradient statistics at random initialization.
\textbf{(a) Gradient Norm Ratio:} The ratio $\|\nabla_{\boldsymbol{z}_h} L\| / \|\nabla_{\boldsymbol{z}_o} L\|$ is consistently close to $1.0$ for both poisoned and clean samples.
\textbf{(b) Cosine Similarity:} The directional alignment between branches exceeds $0.99$ for both data types.
This confirms that the network starts from a strictly fair state, with no initial bias towards the shortcut branch for either data type.}
\label{fig:init_verification}
\end{figure}

\subsubsection{Simplicity Bias and Gradient Dominance}

Recent studies on Simplicity Bias (SB)~\cite{shah2020pitfalls} suggest that neural networks, especially those trained with SGD, exhibit a strong bias towards learning the simplest predictive features (e.g., texture, color, or triggers) over complex semantic patterns.
Consistent with this principle, our shallow shortcut branch $S_h$ represents a lower-capacity model that is inherently more sensitive to simple trigger patterns than the deep backbone $S_o$.
This structural advantage is further amplified by the attention-based gradient focusing derived above. Consequently, the effective parameter sensitivity (Jacobian norm) for poisoned samples satisfies:
\begin{equation}
\left\| \frac{\partial \boldsymbol{f}_h(\boldsymbol{x}_{poisoned})}{\partial \theta_h} \right\| \gg \left\| \frac{\partial \boldsymbol{f}_o(\boldsymbol{x}_{poisoned})}{\partial \theta_o} \right\|.
\label{eq:jacobian_inequality}
\end{equation}

This inequality implies that for poisoned inputs, the optimization step size for the shortcut branch is significantly larger. This creates a ``winner-take-all'' dynamic: the shortcut branch rapidly minimizes the loss on poisoned samples. 
Crucially, this parameter-level dominance manifests in the logit gradients. As $S_h$ begins to fit the trigger pattern faster, the optimizer channels the majority of the error signal through the shortcut path to maximally reduce the loss.
This explains the phenomenon observed in Figure~\ref{fig:grad_dynamics}, where the logit gradient ratio $\|\nabla_{\boldsymbol{z}_h} L\| / \|\nabla_{\boldsymbol{z}_o} L\|$ surges exponentially (exceeding $10^3$ for some attacks, e.g., SIG and CL).
As the total loss $L$ decreases due to the shortcut's fitting, the gradient signal diminishes, leaving the slower-learning deep branch $S_o$ with little incentive—and effectively no gradient signal—to learn the backdoor. This mechanism effectively decouples the backdoor (captured by $S_h$) from the benign knowledge (learned by $S_o$).

However, it is noted that this phenomenon is transient and primarily observable in the early training stages. Without explicit intervention, the deep branch $S_o$ will eventually memorize the backdoor patterns as training progresses toward convergence. Consequently, we design our decoupled training process to explicitly sustain and amplify this initial divergence.

\subsection{Empirical Verification}
\label{sec:verification}

To empirically validate our theoretical derivation, we tracked the gradient dynamics of the logits ($\nabla_{\boldsymbol{z}} L$) during the early training phase (Initialization to Epoch 6) across 8 different backdoor attacks, including invisible (e.g., WaNet), sample-specific (e.g., Dynamic) and clean-label (e.g., SIG, PCBA) attacks.

\paragraph{Balanced Initialization.}
First, we investigate whether the shortcut's dominance stems from an initialized bias. As shown in Figure~\ref{fig:init_verification}, at the random initialization stage, the gradient norm ratio between the shortcut and original branches is consistently $\approx 1.0$, and the cosine similarity exceeds $0.99$ for both benign and poisoned samples. This empirical evidence strictly confirms that the optimization process begins from a fair starting point, where the error signals are unbiased and comparable in magnitude for both branches.

\paragraph{Rapid Divergence and Dominance.}
Starting from this balanced state, we then monitor the evolution of gradient ratios during training. As illustrated in Figure~\ref{fig:grad_dynamics}, a dramatic divergence occurs almost immediately. For poisoned samples (Red lines), the gradient ratio surges exponentially—reaching over $20\times$ for BadNet and over $1000\times$ for SIG and PCBA (note the log scale). In stark contrast, the ratio for clean samples (Blue lines) remains constant at $\approx 1.0$. 

\paragraph{Conclusion.}
These results provide compelling verification of our mechanism: the ``winner-take-all'' phenomenon is not caused by initial signal bias. Instead, it is driven by the \textit{structural sensitivity} of the shortcut branch, which selectively amplifies the consistent gradients from poisoned samples while ignoring the diverse gradients from benign data.

\begin{table*}[t]
\belowrulesep=0pt
\aboverulesep=0pt
\centering

\resizebox{0.8\linewidth}{!}{
    \begin{tabular}{c|c|c|ccccccccc}
    \toprule
    Attack & Metric & No Defense & FP    & NAD   & ABL   & DBD   & CBD   & ASD   & V\&B  & PIPD  & TR (ours)  \\
    \midrule
     \multirow{2}[2]{*}{BadNet} & BA    & 73.14\% & 65.89\% & 61.56\% & 64.94\% & 68.24\% & 65.72\% & 70.18\% & \textbf{75.86\%} & 70.18\% & \underline{75.53\%} \\
                 & ASR   & 100.00\% & 56.75\% & \underline{0.03\%} & 0.89\% & 0.21\% & 0.63\% & 1.96\% & 0.13\% & 0.22\% & \textbf{0.00\%} \\ \hline
          \multirow{2}[2]{*}{Blended} & BA    & 75.93\% & 70.67\% & 64.47\% & 59.24\% & 63.86\% & 62.25\% & 70.23\% & \underline{75.13\%} & 72.51\% & \textbf{75.32\%} \\
                 & ASR   & 99.52\% & 0.88\% & 0.14\% & 1.42\% & \underline{99.70\%} & 40.35\% & 92.84\% & \underline{0.10\%} & 0.39\% & \textbf{0.00\%} \\ \hline
           \multirow{2}[2]{*}{SIG} & BA    & 77.53\% & \underline{73.43\%} & 65.62\% & 59.30\% & 67.83\% & 64.73\% & 70.75\% & 68.63\% & 64.72\% & \textbf{74.42\%} \\
                 & ASR   & 90.90\% & 12.37\% & \underline{0.48\%} & 97.61\% & 75.57\% & 99.07\% & 77.44\% & 3.62\% & 10.92\% & \textbf{0.00\%} \\ \hline
          \multicolumn{1}{c|}{\multirow{2}[2]{*}{WaNet}} & BA    & 74.26\% & \underline{70.19\%} & 68.38\% & 59.46\% & 62.38\% & 62.90\% & 67.01\% & 67.33\% & 69.16\% & \textbf{74.83\%} \\
                 & ASR   & 94.29\% & \underline{4.41\%} & 0.97\% & 1.53\% & 96.07\% & 15.86\% & 42.67\% & 80.58\% & 6.48\% & \textbf{0.00\%} \\ \hline
          \multicolumn{1}{c|}{\multirow{2}[2]{*}{DUBA}} & BA    & 74.33\% & \underline{71.88\%} & 63.92\% & 69.42\% & 65.79\% & 64.24\% & 70.52\% & 70.44\% & \textbf{72.86\%} & 71.72\% \\
                 & ASR   & 95.19\% & 20.53\% & \underline{5.48\%} & 95.20\% & 82.90\% & 10.47\% & 60.08\% & 12.82\% & 9.52\% & \textbf{0.04\%} \\ \hline
          \multirow{2}[2]{*}{\textbf{Average}} & BA    & 75.22\% & 70.41\% & 64.79\% & 62.47\% & 65.62\% & 63.97\% & 69.74\% & \underline{71.48\%} & 69.89\% & \textbf{74.36\%} \\
                 & ASR   & 96.18\% & 18.99\% & \underline{1.42\%} & 39.33\% & 70.89\% & 33.28\% & 55.00\% & 19.45\% & 5.51\% & \textbf{0.01\%} \\
    \bottomrule
    \end{tabular}%
}
\caption{The defense performance on CIFAR-100 using the ResNet-18 backbone. }
\label{defense on CIFAR-100}
\end{table*}

\begin{table*}[tbp]
\belowrulesep=0pt
\aboverulesep=0pt
\centering

\renewcommand\arraystretch{1.1}
\resizebox{0.9\linewidth}{!}{
    \begin{tabular}{c|cc|cc|cc|cc|cc|cc}
    \toprule
    \multicolumn{1}{c|}{\multirow{3}[4]{*}{Poisoning Rate}} & \multicolumn{4}{c|}{ViT-T/16} & \multicolumn{4}{c|}{ViT-S/16} & \multicolumn{4}{c}{ViT-B/16} \\
\cmidrule{2-13}          & \multicolumn{2}{c|}{No Defense} & \multicolumn{2}{c|}{Ours} & \multicolumn{2}{c|}{No Defense} & \multicolumn{2}{c|}{Ours} & \multicolumn{2}{c|}{No Defense} & \multicolumn{2}{c}{Ours} \\
          & BA    & ASR   & BA    & ASR   & BA    & ASR   & BA    & ASR   & BA    & ASR   & BA    & ASR \\
    \midrule
    0.01  & 71.21\% & 0.19\% & 71.05\% & 0.13\% & 79.10\% & 0.15\% & 78.94\% & 0.14\% & 81.90\% & 97.02\% & 81.82\% & 1.26\% \\
    0.1   & 71.08\% & 97.89\% & 69.82\% & 0.76\% & 78.99\% & 98.12\% & 78.89\% & 0.61\% & 81.55\% & 97.27\% & 81.79\% & 0.08\% \\
    \bottomrule
    \end{tabular}%
}
\caption{Defense performance on ImageNet-1K with three ViT models under attacks (\protect\cite{subramanya2024closer}) conducted by fine-tuning on BadNets' poisoned data with varying poisoning rates.}
\label{defense on ImageNet-1K}
\end{table*}

\section{Resistance to Potential Adaptive Attacks}
\label{Appendix：defense against adaptive attacks}
We consider a challenging setting where the attacker knows our defense, evaluating two adaptive attacks in the classical scenario and one in the optimized scenario.

\paragraph{Threat Model for Attackers}
Following existing methods \cite{CBD,ASD}, we assume that attackers have full access to the benign dataset and know the model architecture intended for poisoning. However, they cannot interfere with the training process once the poisoned dataset has been released to the victim users. 

\subsection{Target the Entropy of Poisoned Samples}
\label{B.1}
Our defense leverages the observation that poisoned samples tend to converge faster than benign ones during training~\cite{ABL}. An adaptive attacker may attempt to counter this by intentionally slowing the convergence of poisoned samples, thereby hindering the shortcut branch from capturing backdoor knowledge early on and increasing the likelihood that such knowledge is instead absorbed by the original branch.

Similar to CBD~\cite{CBD}, we describe the adaptive attack as a min-max optimized problem:


\begin{align}
\min_{\theta}\;
\Bigg[
& \sum_{(\boldsymbol{x}, y)\in\mathcal{D}_p}
  \mathcal{L}_{\mathrm{CE}}\!\left(F_\theta(\boldsymbol{x}), y\right) \nonumber\\
& + \sum_{(\boldsymbol{x}_i, y_t)\in\mathcal{D}_m}
    \max_{\delta_i}
    \mathcal{L}_{e}\!\left(F_\theta(\boldsymbol{x}_i+\delta_i)\right)
\Bigg]
\label{adaptive loss}
\end{align}
where $F_\theta$ is a surrogate model with parameter $\theta$. The trigger $\delta_i$ is bounded by $\lVert \delta_i \rVert_\infty < \epsilon$ to make it invisible to visual inspection. We adopt projected gradient descent (PGD) \cite{PGD} to optimize the trigger pattern $\delta_i$ for each poisoned sample step by step:

\begin{equation}
    \boldsymbol{x}_{t+1} = \Pi_{\epsilon}(\boldsymbol{x}_t + \beta \cdot \nabla_{\boldsymbol{x}} \mathcal{L}_e(F_\theta(\boldsymbol{x}_t))
\end{equation}
where $t$ is the current step ($M$ steps in total), $\beta$ is the step size, $\mathcal{L}_{e}$ refers to the entropy introduced in Equation \ref{entropy}, $\Pi_{\epsilon}$ clips the updated images to the range $[0, 255]$ and restricts the cumulative perturbation $\delta_i$ within the $\epsilon$ bound.

\paragraph{Experimental Settings.}
We conduct the adaptive attack on the CIFAR-10 dataset using a ResNet-18 network. Following previous work on adversarial attacks, we set $\epsilon = 8/255$ to ensure the trigger remains imperceptible to the human visual system. We use SGD to solve the above optimization problem over 10 epochs with $\beta=0.001$ and $M = 3$. The initial learning rate is set to 0.01 and is reduced by a factor of 10 halfway through the training.

\paragraph{Results.}
The adaptive attack successfully injects a backdoor into the model, achieving a BA of 88.23\% and an ASR of 99.99\%. However, our defense reduces its ASR to 0\% and improves the BA to 94.49\%. This suggests that the optimized trigger patterns become less effective when the model is retrained with randomly initialized parameters. In other words, the optimized triggers exhibit limited transferability across different model instances.

\subsection{Target the Shortcut Learning Behavior}
Our defense is based on the assumption that poisoned samples are more likely to be learned by a shortcut branch during the early stages of training. To circumvent this, an adaptive attacker may target our network architecture by intentionally suppressing the learning of poisoned samples in the shortcut branch. The corresponding optimized attack can be formulated as follows:


\begin{align}
\min_{\theta}\;
\Bigg[
& \sum_{(\boldsymbol{x}, y)\in\mathcal{D}_p}
  \mathcal{L}_{\mathrm{CE}}\!\left(P(\boldsymbol{x}), y\right) \nonumber \\
& + \sum_{(\boldsymbol{x}_i, y_t)\in\mathcal{D}_m}
    \max_{\delta_i}
    \mathcal{L}_{\mathrm{CE}}\!\left(P_h(\boldsymbol{x}_i+\delta_i), y_t\right)
\Bigg]
\end{align}

\paragraph{Experimental Settings.}
Same as section~\ref{B.1}.

\paragraph{Results.}
The adaptive attack results in a BA/ASR of 81.04\%/99.84\% for the original branch and 52.59\%/55.59\% for the shortcut branch, indicating that the learning of poisoned samples in the shortcut branch has been effectively suppressed. However, after applying our defense, the model achieves a BA of 94.16\% and an ASR of just 0.01\%. Notably, during the defense process, the poisoned samples are still initially absorbed by the shortcut branch, suggesting that the optimized triggers exhibit poor transferability across training instances.

\subsection{Targeting the Distinguishability Between Poisoned and Benign Samples}
During training, our defense decouples the features of poisoned and benign samples and guides them to be learned through different branches. An adaptive attacker may attempt to construct poisoned samples whose features closely resemble those of benign samples, thereby weakening the feature-level separation induced by our defense. Such an adaptive attack can be formulated as:
\begin{align}
\min_{\theta}\;
\Bigg[
& \sum_{(x,y)\in D_p} L_{\mathrm{CE}}(P(x),y) \nonumber \\
& + \sum_{(x_i,y_t)\in D_m}
    \max_{\delta_i}
    \langle S(x_i+\delta_i),\, S(x_i)\rangle
\Bigg]
\end{align}
where $S(\cdot)$ denotes the feature extractor.

\paragraph{Experimental Settings.}
We conduct this adaptive attack on the CIFAR-10 dataset using a WRN-16-1 network. All other experimental settings remain the same as those in Section~\ref{B.1}.

\paragraph{Results.}
The adaptive attack achieves a BA/ASR of 75.58\%/80.84\%. Under this setting, our defense reduces the ASR to 1.90\% while improving the BA to 90.03\%. Compared with the previous two adaptive attacks, this attack leaves more backdoor knowledge in the original branch. Nevertheless, our defense remains effective, as the strong stealthiness enforced by the attack-specific loss is difficult to transfer when the model is trained without such objectives.

\section{Evaluation on Additional Datasets and Model Architectures}
\label{Appendix: Evaluation on Additional Datasets and Model Architectures}

\subsection{Effectiveness on CIFAR-100}
\label{Appendix: defense on CIFAR-100}
To evaluate the effectiveness of our method with a larger number of classes, we analyze its defense performance on the CIFAR-100 dataset using the ResNet-18 backbone, as shown in Table~\ref{defense on CIFAR-100}. The results clearly demonstrate that our TR reduces the attack success rate of all attacks to 0\% while achieving either the best or second-best benign accuracy. This superior performance stems from our decoupling strategy, which channels all backdoor knowledge into the shortcut branch, enabling the original network to focus on learning benign information more effectively. Consequently, our method even outperforms the baselines (\ie\ No Defense) in benign accuracy against BadNets and WaNet. 

Compared to CIFAR-10, we observe that the shortcut branch captures backdoor knowledge more quickly on CIFAR-100. This is likely because the increased inter-class differences and complexity make the shared trigger patterns more distinctive, allowing the shortcut branch to learn them more easily. This suggests that our method has the potential to scale effectively to datasets with more classes.

\begin{table*}[t]
\belowrulesep=0pt
\aboverulesep=0pt
\centering
\renewcommand\arraystretch{1.3}
\Huge
\resizebox{\linewidth}{!}{
    \begin{tabular}{c|c|cc|cc|cc|cc|cc|cc|cc|cc|cc|cc|cc}
    \toprule
    \multirow{2}[4]{*}{Model} & \multirow{2}[4]{*}{Defense} & \multicolumn{2}{c|}{BadNet} & \multicolumn{2}{c|}{Blended} & \multicolumn{2}{c|}{WaNet} & \multicolumn{2}{c|}{Dynamic} & \multicolumn{2}{c|}{DataFree} & \multicolumn{2}{c|}{DUBA} & \multicolumn{2}{c|}{SIG} & \multicolumn{2}{c|}{CL} & \multicolumn{2}{c|}{Narcissus} & \multicolumn{2}{c|}{PCBA} & \multicolumn{2}{c}{\textbf{Average}} \\
\cmidrule{3-24}          &       & BA    & ASR   & BA    & ASR   & BA    & ASR   & BA    & ASR   & BA    & ASR   & BA    & ASR   & BA    & ASR   & BA    & ASR   & BA    & ASR   & BA    & ASR   & BA    & ASR \\
    \midrule
    \multicolumn{1}{c|}{\multirow{12}[4]{*}{ResNet-18}} & No Defense & 94.36  & 100.00  & 94.57  & 99.76  & 94.21  & 95.84  & 94.66  & 95.18  & 94.66  & 100.00  & 94.72  & 95.86  & 95.09  & 64.79  & 95.03  & 94.73  & 94.21  & 99.36  & 93.49  & 99.90  & 94.50  & 94.54  \\
\cmidrule{2-24}          & FP    & 92.72  & 1.92  & 92.49  & 10.61  & \underline{93.27 } & 0.90  & 92.47  & 11.73  & 92.89  & 1.54  & 91.85  & 19.37  & 93.55  & 37.98  & 93.03  & 6.39  & 88.61  & 50.43  & 85.72  & 65.38  & 91.66  & 20.63  \\
          & NAD   & 88.83  & 1.83  & 87.72  & 1.38  & 91.00  & 0.99  & 88.30  & 2.32  & 88.41  & 1.33  & 85.33  & 27.01  & 90.42  & 3.52  & 90.19  & 2.60  & 88.36  & 32.84  & 87.83  & 29.23  & 88.64  & 10.31  \\
          & ABL   & 90.47  & 0.70  & 90.17  & 7.04  & 80.78  & 19.93  & 83.07  & 94.21  & 89.70  & 3.92  & 80.64  & 43.90  & 92.34  & 5.61  & 86.50  & 2.68  & 72.53  & 41.73  & 70.34  & 62.45  & 83.65  & 28.22  \\
          & DBD   & 92.00  & 3.06  & 91.62  & 3.78  & 89.13  & \textbf{0.11 } & 90.98  & 17.56  & 89.27  & 5.86  & 89.73  & 3.92  & 89.42  & 1.61  & 93.24  & 14.88  & 76.25  & 73.58  & 78.60  & 87.14  & 88.02  & 21.15  \\
          & CBD   & 85.87  & 2.13  & 83.40  & 15.27  & 88.83  & 79.71  & 88.54  & 33.84  & 85.29  & 2.45  & 81.66  & 96.58  & 85.84  & 16.71  & 83.33  & 1.72  & 81.29  & 24.83  & 82.61  & 74.06  & 84.67  & 34.73  \\
          & ASD   & 93.44  & 1.09  & 93.18  & 3.62  & 93.10  & 2.41  & 92.32  & 6.92  & 93.30  & 2.31  & 89.03  & \underline{0.27 } & 92.13  & 1.07  & 93.71  & 3.03  & 87.92  & 92.23  & 85.16  & 92.29  & 91.33  & 20.52  \\
          & V\&B & \underline{93.96 } & 0.62  & 93.67  & 0.53  & \textbf{94.05 } & \underline{0.54 } & \textbf{93.91 } & 1.13  & \underline{93.86 } & 0.56  & \textbf{94.27 } & \textbf{0.01 } & \underline{94.08 } & 0.17  & 93.98  & 0.64  & 89.87  & 25.62  & 87.58  & 13.37  & \underline{92.92 } & 4.32  \\
          & PIPD  & 93.28  & 0.55  & 93.22  & 0.78  & 93.27  & 2.80  & 92.11  & 8.03  & 90.62  & 3.58  & 89.37  & 5.30  & 93.79  & \underline{0.03 } & 93.62  & 1.24  & 89.35  & 16.29  & 83.84  & 14.52  & 91.25  & 5.31  \\
          & PDB   & 81.36  & 0.74  & 81.69  & 1.66  & 80.95  & 1.90  & 81.85  & 2.27  & 81.37  & 0.87  & 82.62  & 2.49  & 80.75  & 0.68  & 81.57  & 1.73  & 82.47  & 1.85  & 82.41  & 0.81  & 81.70  & \underline{1.50 } \\
          & ESTI  & 93.75  & \textbf{0.00 } & \underline{93.82 } & \textbf{0.00 } & 78.98  & 1.96  & 92.95  & \textbf{0.00 } & 92.53  & \textbf{0.00 } & \underline{93.36 } & 23.88  & 90.50  & 92.32  & \underline{94.12 } & \textbf{0.00 } & \underline{92.25 } & \textbf{0.00 } & \underline{89.96 } & \textbf{0.00 } & 91.22  & 11.82  \\
          & TR (Ours) & \textbf{94.11 } & \textbf{0.00 } & \textbf{94.07 } & \textbf{0.00 } & 92.82  & 1.86  & \underline{93.41 } & \underline{0.76 } & \textbf{94.21 } & \textbf{0.00 } & 91.18  & 0.77  & \textbf{94.70 } & \textbf{0.00 } & \textbf{94.62 } & \underline{0.74 } & \textbf{93.52 } & \underline{0.27 } & \textbf{91.78 } & \textbf{0.00 } & \textbf{93.44 } & \textbf{0.44 } \\
    \midrule
    \multicolumn{1}{c|}{\multirow{4}{*}{\begin{tabular}{c}
PreAct \\
ResNet-18
\end{tabular}}} & No Defense & 93.49  & 100.00  & 94.39  & 99.61  & 92.44  & 97.36  & 92.98  & 92.74  & 93.55  & 99.83  & 93.08  & 96.70  & 93.44  & 78.36  & 93.24  & 45.89  & 93.81  & 95.49  & 93.54  & 98.11  & 93.40  & 90.41  \\
\cmidrule{2-24}          & PDB   & 90.16  & 0.42  & 92.63  & 0.20  & \underline{90.68 } & 2.18  & 90.73  & 1.89  & 91.55  & 0.80  & 87.08  & 5.72  & \underline{92.79 } & 0.12  & 92.83  & \underline{0.27 } & 91.72  & 1.79  & 91.46  & 2.57  & \underline{91.16 } & 1.60  \\
          & ESTI  & \underline{93.92 } & \textbf{0.00 } & \textbf{93.96 } & \textbf{0.00 } & 84.53  & \underline{1.24 } & \underline{91.50 } & \textbf{0.42 } & \underline{93.95 } & \textbf{0.00 } & \underline{89.31 } & \textbf{2.08 } & 44.23  & \textbf{0.00 } & \underline{93.77 } & \textbf{0.00 } & \underline{93.72 } & \textbf{0.92 } & \textbf{93.05 } & \textbf{0.00 } & 87.19  & \textbf{0.47 } \\
          & TR (Ours) & \textbf{94.43 } & \textbf{0.00 } & \underline{93.43 } & \underline{0.02 } & \textbf{91.09 } & \textbf{0.90 } & \textbf{93.69 } & \underline{0.59 } & \textbf{94.69 } & \textbf{0.00 } & \textbf{91.34 } & \underline{2.11 } & \textbf{94.02 } & \textbf{0.00 } & \textbf{93.80 } & 0.57  & \textbf{94.33 } & \underline{1.03 } & \underline{92.76 } & \underline{0.16 } & \textbf{93.36 } & \underline{0.54 } \\
    \bottomrule
    \end{tabular}%
}
\caption{The defense performance on CIFAR-10 using the ResNet-18 and PreActResNet-18 backbone.}
\label{defense on CIFAR-10 with ResNet-18}
\end{table*}

\begin{table*}[tbp]
\belowrulesep=0pt
\aboverulesep=0pt
\centering
\renewcommand\arraystretch{1.1}
\Large
\resizebox{0.9\linewidth}{!}{
    \begin{tabular}{c|cc|cc|cc|cc|cc|cc|cc|cc}
    \toprule
    \multirow{2}[4]{*}{Attack} & \multicolumn{2}{c|}{BadNets} & \multicolumn{2}{c|}{Blend} & \multicolumn{2}{c|}{WaNet} & \multicolumn{2}{c|}{DataFree} & \multicolumn{2}{c|}{DUBA} & \multicolumn{2}{c|}{SIG} & \multicolumn{2}{c|}{Narcissus} & \multicolumn{2}{c}{\textbf{Average}} \\
\cmidrule{2-17}          & BA    & ASR   & BA    & ASR   & BA    & ASR   & BA    & ASR   & BA    & ASR   & BA    & ASR   & BA    & ASR   & BA    & ASR \\
    \midrule
    No Defense & 96.39 & 100.00 & 96.44 & 100.00 & 94.61 & 89.41 & 98.65 & 100.00 & 96.84 & 98.64 & 96.54 & 100.00 & 98.80 & 50.96 & 96.90 & 91.29 \\
    \midrule
    FP    & 94.91 & 0.19  &\underline{95.31} & 12.51 &\underline{95.70} & 1.07  & 92.82 & 0.73  & 93.25 & 10.38 &\underline{94.80} & 10.35 & 94.35 & 14.36 & 94.45 & 7.08 \\
    NAD   &\underline{95.83} & 0.37  & 94.96 &\underline{7.46} & 93.34 &\underline{0.60} &\underline{94.46} & 0.04  &\underline{94.23} & 3.81  & 94.29 & 0.64  &\underline{94.62} & 5.98  &\underline{94.53} &\underline{2.70} \\
    ABL   & 91.66 & 6.00  & 86.27 & 39.98 & 93.60 & 76.11 & 91.63 & 0.03  & 90.64 & 51.77 & 84.26 & 100.00 & 85.27 & 92.41 & 89.05 & 52.33 \\
    ESTI  & 94.62 & \textbf{0.00} & 90.68 & 15.93 & 48.49 & 31.56 & 94.00 & \textbf{0.00} & 91.15 & \textbf{0.00} & 82.88 & \textbf{0.00} & 84.18 & \textbf{0.00} & 83.71 & 6.78 \\
    TR(Ours) & \textbf{96.35} &\underline{0.01} & \textbf{95.92} & \textbf{0.00} & \textbf{96.25} & \textbf{0.00} & \textbf{96.47} & \textbf{0.00} & \textbf{95.87} &\underline{0.38} & \textbf{95.99} &\underline{0.08} & \textbf{96.14} & \textbf{0.00} & \textbf{96.14} & \textbf{0.07} \\
    \bottomrule
    \end{tabular}%
    
}
\caption{The defense performance on GTSRB using the WRN-16-1 backbone.}
\label{defense on GTSRB with WRN-16-1}
\end{table*}

\subsection{Effectiveness under More Dataset-Network Combinations}
\label{Appendix: Effectiveness across different network architectures}
To better demonstrate the effectiveness of our method across different network architectures, we compare it with other defenses on CIFAR-10 using ResNet-18 and PreActResNet-18 in Table~\ref{defense on CIFAR-10 with ResNet-18}, and on GTSRB using WRN-16-1 in Table~\ref{defense on GTSRB with WRN-16-1}. Our method achieves the best or second-best performance in most cases, particularly on the GTSRB dataset. Although ESTI achieves a slightly lower average ASR with PreActResNet-18, our approach yields a substantially higher average BA. Consistent performance across three network architectures and two datasets demonstrates the strong generalization capability of our method. 

\begin{table}[tbp]
\belowrulesep=0pt
\aboverulesep=0pt
\centering
\caption{Comparison with defense focused on fine-tuning attacks on CIFAR-10 using the VGG-16 backbone.}
\resizebox{0.99\linewidth}{!}{
    \begin{tabular}{c|cc|cc}
    \toprule
    \multirow{2}[4]{*}{Method} & \multicolumn{2}{c|}{White Square} & \multicolumn{2}{c}{Black Line} \\
\cmidrule{2-5}          & BA    & ASR   & BA    & ASR \\
    \midrule
    No Defense & 91.33\% & 100\% & 91.28\% & 100\% \\
    \midrule
    Tang~\cite{setting}     & 92.20\% & 8.81\% & 92.23\% & 10.81\% \\
    Ours & \textbf{92.30\%} & \textbf{0.01\%} & \textbf{92.62\%} & \textbf{0.14\%} \\
    \bottomrule
    \end{tabular}
}
\label{Comparison with NLP-focused Defense on CIFAR-10}
\end{table}

\subsection{Effectiveness on ViTs and ImageNet-1k}
\label{Appendix: Deit and ImageNet}
In Table~\ref{defense on ImageNet-1K}, we attack three different-sized ViT models using the BadNets attack, following the approach in \cite{subramanya2024closer}. All three models consist of 12 transformer blocks but differ in feature dimensions and the number of attention heads. Specifically, ViT-T/16, ViT-S/16, and ViT-B/16 have feature dimensions of 192, 384, and 768, respectively, with 3, 6, and 12 attention heads. 

For defense, we introduce a honeypot shortcut comprising 6 transformer blocks instead of a Conv-based shortcut to better align with the original feature space. An ablation study on shortcut size is provided in Table~\ref{Ablation study of the shortcut size for ViT}. Since the default poisoning rate of 0.01 fails to attack ViT-T/16 and ViT-S/16, we increase it to 0.1 for a stronger attack.

Similar to NLP backdoor attacks, these attacks are implemented by fine-tuning a pre-trained clean model with poisoned data, a common approach when targeting transformers. In this scenario, benign and backdoor knowledge are naturally decoupled from the start. The benign samples are inherently learned by the clean original branch, while the poisoned samples are rapidly captured by our shortcut branch. Consequently, our defense effectively reduces the attack success rate to below or close to 1\%, with only a minimal impact on benign accuracy. We provide another defense results against fine-tuned attacks to CNN in Appendix~\ref{Appendix: Comparison with NLP-focused Defense}.


\subsection{Comparison with Defenses Focused on Fine-Tuning Attacks (VGG-16)}
\label{Appendix: Comparison with NLP-focused Defense}

Tang~\cite{setting} focus on NLP backdoors, proposing trapping backdoors within shallow layers and extend their method to image classification. Following their setup, we fine-tune a clean VGG-16 network pretrained on ImageNet on CIFAR-10, using a $3 \times 3$ white square and a black line (3 pixels wide) as triggers. The white square is placed at the bottom-right corner of the image, while the black line is positioned at the bottom.

As shown in ~\ref{Comparison with NLP-focused Defense on CIFAR-10}, our defense remains superior to \cite{setting} in the fine-tuned attack scenario, primarily due to a more effective honeypot design. Although \cite{setting} employs loss functions to prevent deep layers from learning backdoors captured by shallow layers, the backdoor knowledge in shallow layers can still influence final predictions.  In contrast, our method introduces a separate parallel branch to absorb backdoor knowledge, ensuring it does not affect the original network.

\begin{table*}[t]
  \belowrulesep=0pt
  \aboverulesep=0pt
  \caption{Class-wise BA results on CIFAR-10 and unbalanced GTSRB. In GTSRB, the selected classes are rare, each comprising less than 0.8\% of the training data.}
  \centering
  \label{Table: class-wise performance}
  \renewcommand\arraystretch{1.1}
  \resizebox{0.9\linewidth}{!}{
    \begin{tabular}{c|c|cccccccccc}
    \toprule
    \multirow{3}[2]{*}{CIFAR-10} & Class & 0     & 1     & 2     & 3     & 4     & 5     & 6     & 7     & 8     & 9 \\
    \cmidrule{2-12}
          & Clean & 90.80\% & 96.00\% & 87.10\% & 82.90\% & 92.40\% & 86.90\% & 93.10\% & 93.00\% & 94.40\% & 94.30\% \\
          & Ours  & 84.40\% & 95.70\% & 86.70\% & 80.10\% & 92.70\% & 85.50\% & 92.00\% & 92.70\% & 96.40\% & 94.00\% \\
    \midrule
    \multirow{3}[2]{*}{GTSRB} & Class & 0     & 19    & 24    & 27    & 29    & 32    & 37    & 41    & 42    & Mean \\
    \cmidrule{2-12}
          & Clean & 100.00\% & 86.67\% & 98.89\% & 76.67\% & 96.67\% & 100.00\% & 43.33\% & 98.33\% & 100.00\% & 88.95\% \\
          & Ours  & 100.00\% & 96.67\% & 100.00\% & 69.67\% & 97.78\% & 100.00\% & 38.33\% & 93.33\% & 100.00\% & 88.53\% \\
    \bottomrule
    \end{tabular}%
  }
\end{table*}

\section{More Experiments and Analysis}

\subsection{Robustness to Different Poisoning Rates}
\label{Appendix: poisoning rate}

\begin{figure}[tbp]
\begin{center}
\includegraphics[width=0.7\linewidth]{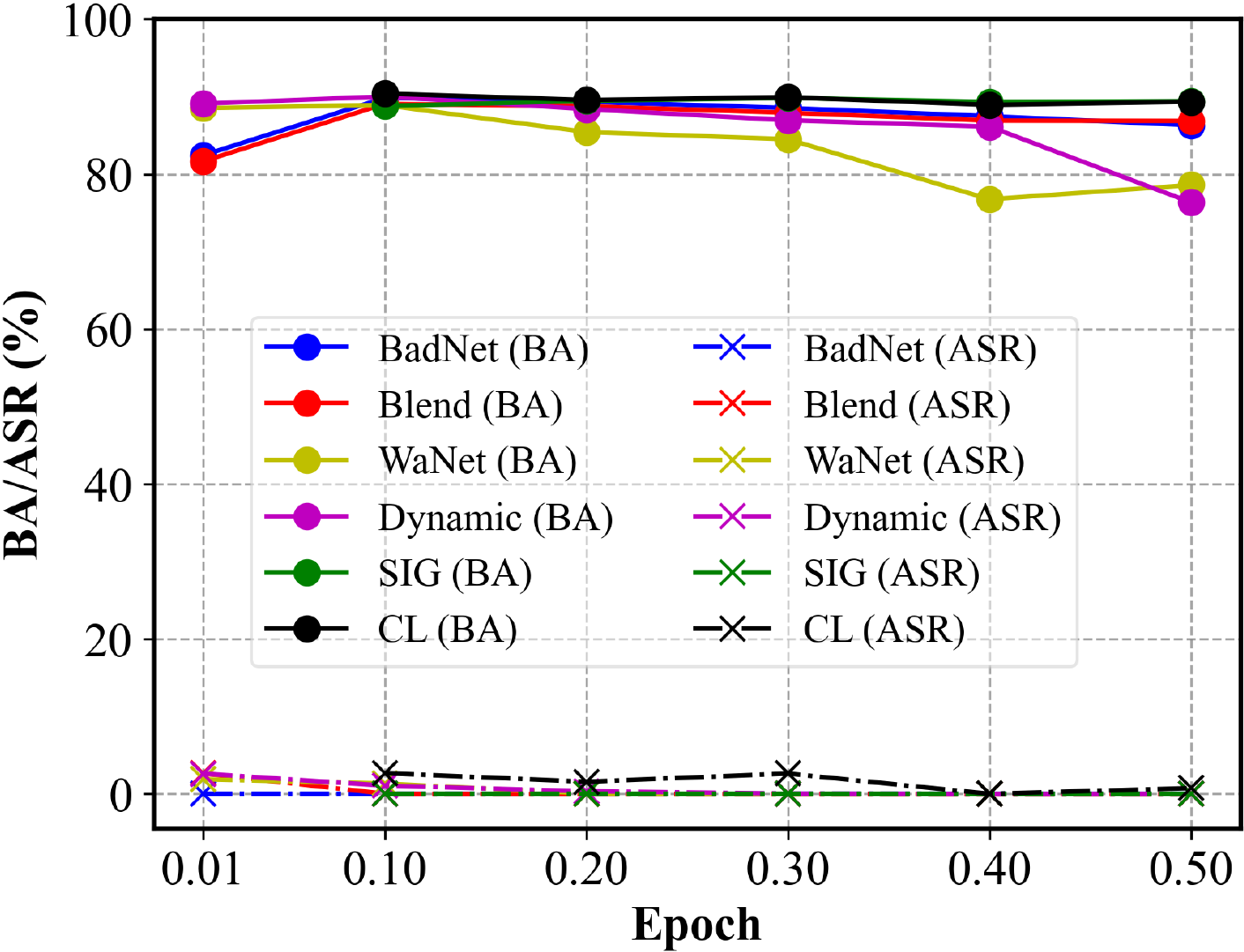}
\end{center}
   \caption{The BA and ASR under different poisoning rates.}
\label{fig: poisoning rate}
\end{figure}

We verify the effectiveness of our method across poisoning rates ranging from 0.01 to 0.5, and the results presented in Figure~\ref{fig: poisoning rate}. In this subsection, we redefine the poisoning rate for clean-label attacks as the proportion of poisoned samples within the target-class samples. Thus, the poisoning rates for SIG and CL attacks are 0.5 in Table~\ref{defense against classical attack}, since their attack success rates only reach 48.18\% and 1.78\%, respectively, when the poisoning rate is 0.1. For the same reason, we omit evaluations against these attacks under the extremely low poisoning rate of 0.01.

The results show that our method can reduce the attack success rate to nearly 0\% in most cases, while maintaining satisfactory benign accuracy. As the poisoning rate increases, the shortcut branch is able to capture backdoor knowledge more quickly and effectively, providing stronger protection to the original network from learning the backdoor knowledge. This is why our method performs well even with a poisoning rate as high as 0.5. 

We observe a drop in the benign accuracy of WaNet and Dynamic when the poisoning rate exceeds 0.3 and 0.4, respectively.  This occurs because both WaNet and Dynamic employ a noisy training mode, where an additional 20\% and 10\% of benign samples are injected with triggers but maintain the ground-truth labels. Consequently, when the poisoning rate reaches 0.5, only 30\% and 40\% of the samples remain benign in WaNet and Dynamic, respectively.

We also note that the benign accuracy for BadNets and Blend is relatively lower under the extremely low poisoning rate of 0.01. This is due to their weak attack strength, which makes the backdoor knowledge less distinguishable from certain easily learned benign knowledge. Consequently, the shortcut branch captures a small proportion of benign knowledge during training, resulting in a relatively lower benign accuracy for the original network. Nevertheless, our method remains highly effective in inhibiting backdoor injection by absorbing most of the backdoor knowledge.

\subsection{Class-wise Benign Performance}
\label{Appendix: Class-wise Benign Performance}
Although the BA drops reported in Table~\ref{defense against classical attack} appear small, we further present detailed class-wise BA for models trained on clean data and models trained on poisoned data with our defense in Table~\ref{Table: class-wise performance}, to more clearly demonstrate the effectiveness of our approach. Across both datasets, our defense achieves class-wise BA comparable to that of clean models, with only minor decreases in certain cases.

For the CIFAR-10 dataset, although the shortcut branch may initially absorb some easy benign samples, our decoupling process effectively transfers most benign knowledge back to the original branch. For the imbalanced GTSRB dataset, due to the shortcut branch’s limited capacity, it fails to capture rare classes and, guided by the decoupling losses, tends to misassign them to other categories. Consequently, the original branch primarily learns to handle these rare-class samples, resulting in minimal performance degradation even under imbalance. 

In summary, our method preserves most easy benign knowledge while also ensuring that rare benign knowledge is not neglected.

\begin{figure}[tbp]
\begin{center}
\includegraphics[width=0.99\linewidth]{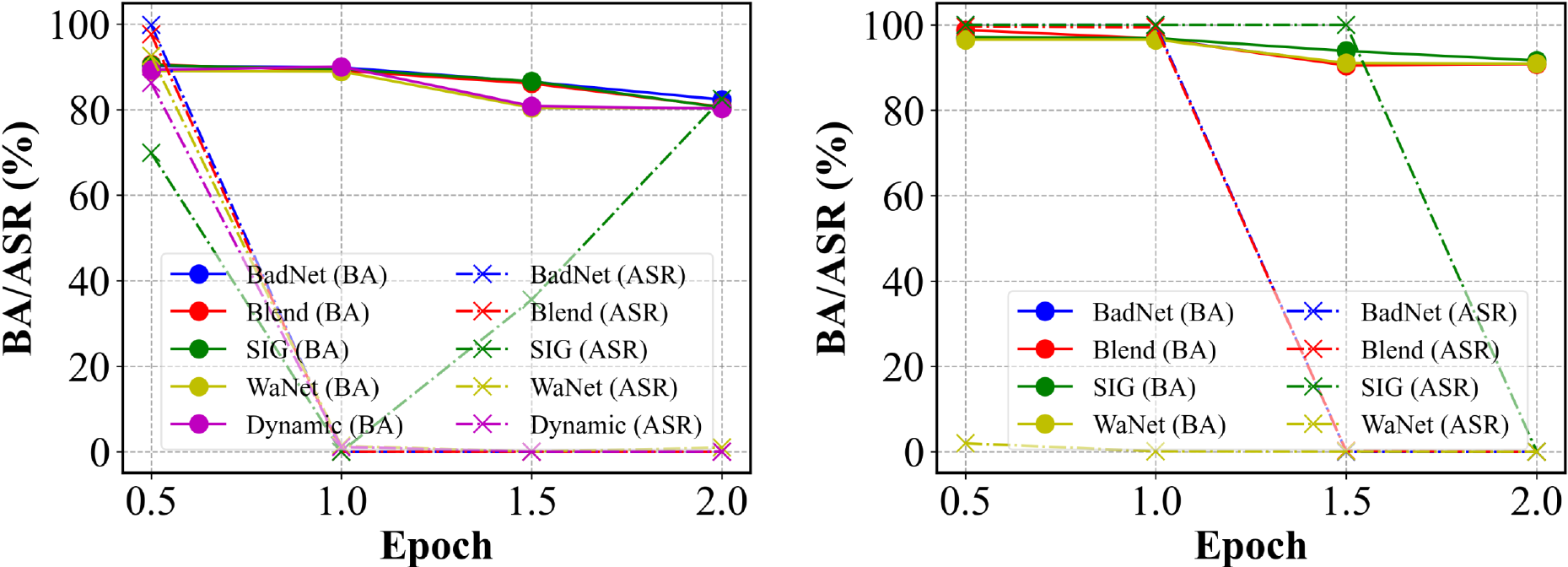}
\end{center}
   \caption{Ablation study of the decoupling weight $\alpha$ on CIFAR-10 (left) and GTSRB (right).}
\label{fig: ablation study alpha}
\end{figure}

\subsection{Selection of Weight $\alpha$}
\label{Appendix: weight alpha}

As Figure~\ref{fig: ablation study alpha} shows, we adjust the decoupling weight $\alpha$ from 0.5 to 2 on CIFAR-10 (left subfigure) and GTSRB (right subfigure) to study its impact on model performance. When $\alpha$ is small, the attack success rate remains high on both datasets because weak decoupling fails to prevent the original network from learning backdoor knowledge, resulting in high attack success rates for both branches. Conversely, when $\alpha$ is sufficiently large (\eg\ 1 for CIFAR10 and 2 for GTSRB), the success rate of all attacks drops to nearly 0. However, an excessively large $\alpha$ (\eg\ 1.5 and 2 for both datasets) hinders the original network from effectively learning benign knowledge initially captured by the shortcut branch, leading to a decline in benign accuracy. For clean-label attack SIG, backdoor knowledge is more vulnerable than in dirty-label attacks due to competition between the trigger pattern and the target-label ground-truth pattern. When $\alpha$ becomes too large, the loss $L_h$ struggles to retain the backdoor knowledge within the shortcut branch, causing it to be disrupted and absorbed by the original network. Consequently, the attack success rate for SIG on CIFAR-10 increases when $\alpha$ exceeds 1. To strike a balance between benign accuracy and attack success rate, we set $\alpha$ to 1 for CIFAR-10.

For GTSRB, which includes a larger number of similar classes (traffic signs), decoupling is generally more challenging, and defense failures persist until $\alpha$ reaches 2. However, setting $\alpha = 2$ throughout training compromises benign accuracy. To address this, we linearly decrease $\alpha$ from 2 to 1 during the first 50 epochs. The initially large $\alpha$ effectively prevents the original network from learning backdoor knowledge in early epochs. In the subsequent epochs, a smaller $\alpha$ suffices to maintain a low attack success rate, as the learning directions of the two branches have been roughly determined. Simultaneously, the gradual decrease of $\alpha$ gradually relaxes the control over the benign knowledge captured by the shortcut branch, allowing it to be learned by the original network.


\begin{table}[tbp]
\belowrulesep=0pt
\aboverulesep=0pt
\centering

\resizebox{0.99\linewidth}{!}{
    \begin{tabular}{c|c|cccc}
    \toprule
    Attacks & Metric & Small & W/O attention & Base  & Large \\
    \midrule
    \multirow{2}[2]{*}{BadNets} & BA    & 89.42\% & 89.68\% & 89.83\% & 90.49\% \\
          & ASR   & 0.00\% & 0.00\% & 0.00\% & 0.00\% \\
    \midrule
    \multirow{2}[2]{*}{Blend} & BA    & 89.75\% & 89.21\% & 89.06\% & 89.78\% \\
          & ASR   & 0.00\% & 0.00\% & 0.01\% & 0.01\% \\
    \midrule
    \multirow{2}[2]{*}{WaNet} & BA    & 88.00\% & 88.79\% & 88.90\% & 88.54\% \\
          & ASR   & 81.71\% & 81.41\% & 1.33\% & 80.41\% \\
    \midrule
    \multirow{2}[2]{*}{Dynamic} & BA    & 89.57\% & 89.13\% & 90.01\% & 89.12\% \\
          & ASR   & 86.34\% & 88.08\% & 1.06\% & 65.16\% \\
    \midrule
    \multirow{2}[2]{*}{CL} & BA    & 89.22\% & 89.64\% & 89.33\% & 89.17\% \\
          & ASR   & 1.01\% & 2.73\% & 0.74\% & 1.07\% \\
    \bottomrule
    \end{tabular}%
}
\caption{Ablation study of the shortcut size for CNN.}
\label{Ablation study of the shortcut size for CNN} 
\end{table}

\begin{figure}[t]
\begin{center}
\includegraphics[width=0.7\linewidth]{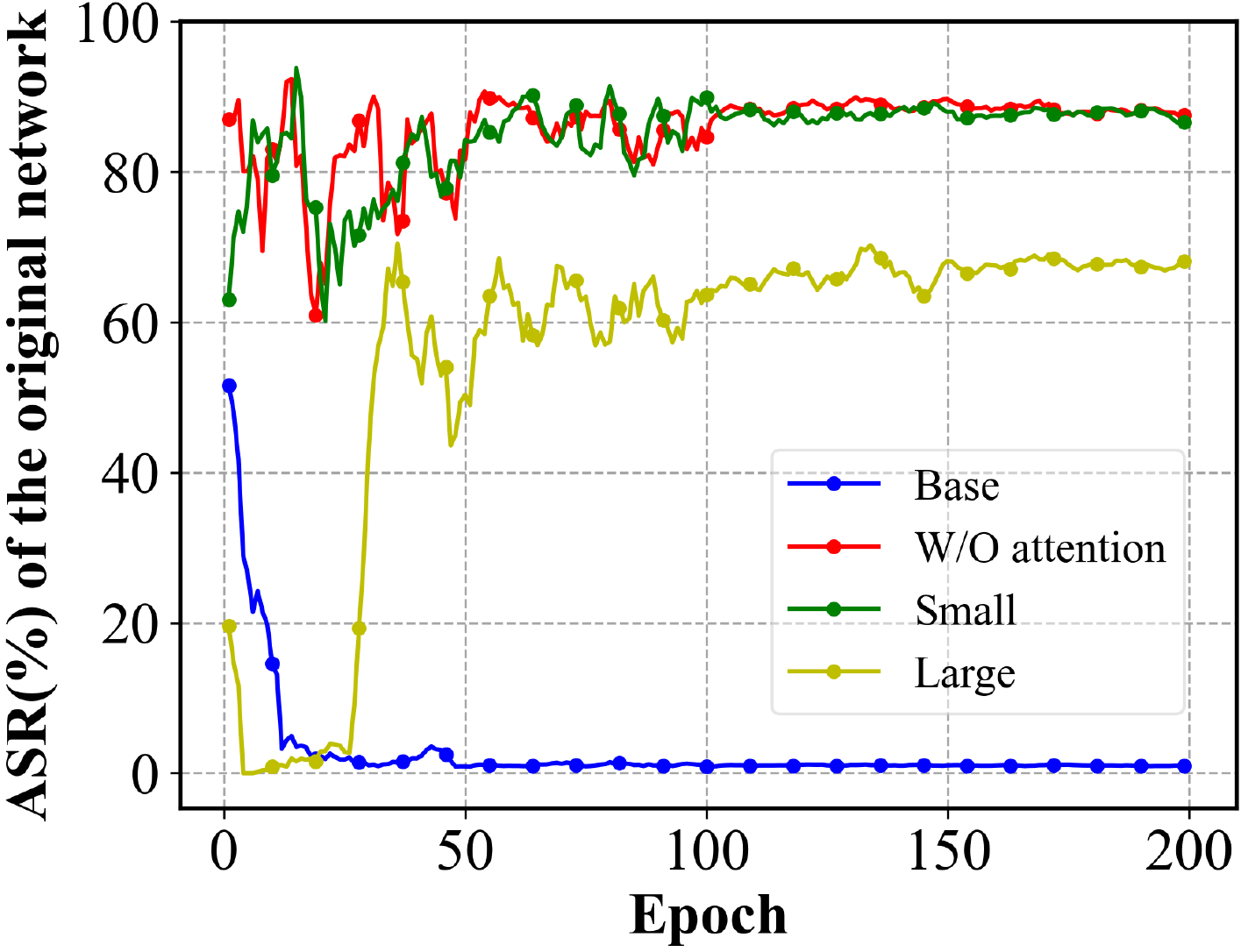}
\end{center}
   \caption{The ASR of our method against the Dynamic attack under different shortcut configurations.}
\label{fig: Shortcut}
\end{figure}

\begin{table}[t]
\belowrulesep=0pt
\aboverulesep=0pt
\centering
\resizebox{0.6\linewidth}{!}{
    \begin{tabular}{c|cc}
    \toprule
    Shortcut Size & BA    & ASR \\
    \midrule
    1 block & 81.84\% & 2.69\% \\
    3 blocks & 81.81\% & 0.08\% \\
    6 blocks & 81.79\% & 0.08\% \\
    9 blocks & 81.77\% & 0.07\% \\
    \bottomrule
    \end{tabular}%
}
\caption{Ablation study of the shortcut size for ViT.}
\label{Ablation study of the shortcut size for ViT}
\end{table}

\begin{table*}[t]
\belowrulesep=0pt
\aboverulesep=0pt
\centering
\renewcommand\arraystretch{1.2}
\resizebox{\textwidth}{!}{%
        \begin{tabular}{l|cccccccccc|c|c}
            \toprule
            Branch & Class 0 & Class 1 & Class 2 & Class 3 & Class 4 & Class 5 & Class 6 & Class 7 & Class 8 & Class 9 & \textbf{Entropy} & \textbf{Weight} \\
            \midrule
            Shortcut & \textbf{9.9999e-01} & 3.60e-09 & 4.14e-08 & 1.12e-05 & 1.30e-09 & 2.13e-09 & 2.28e-08 & 5.81e-09 & 1.98e-07 & 6.71e-09 & 0.0001 & $w_h \approx 1$ \\
            Original & 2.49e-17 & 3.74e-17 & 3.18e-19 & 1.70e-18 & 1.36e-19 & 1.68e-19 & 7.87e-19 & 6.27e-19 & \textbf{1.0000e+00} & 1.15e-17 & 1.79e-15 & $w_o \approx 0$ \\
            \bottomrule
        \end{tabular}%
    }
\caption{The softmax prediction distributions of both branches for a BadNet poisoned sample, along with their corresponding entropy and weight values. Here, Class 0 is the target class. The shortcut predicts this poisoned sample as the target label with extremely high confidence, as expected. However, the original branch also makes a highly confident prediction into a non-target, incorrect class (Class 8). This results in an even lower entropy. Consequently, according to our entropy-based weight assignment, the original branch receives a very small $w_o$, while the shortcut dominates the final prediction with $w_h \approx 1$.}
\label{Table: case study}
\end{table*}

\subsection{Impact of the Shortcut Size}
\label{Appendix: Impact of the Shortcut Size}



To illustrate the impact of shortcut size on our method’s performance, we compare our base design with three variants in Table~\ref{Ablation study of the shortcut size for CNN}. The experiments are conducted on CIFAR-10 using the WRN-16-1 network. Following the same settings as in Table~\ref{defense against classical attack}, our \emph{Base} shortcut consists of two \emph{Conv-Attention-BN-ReLU} blocks. Additionally, we define a \emph{Small} shortcut with a single block and a \emph{Large} shortcut with three blocks. Furthermore, we introduce a \emph{W/O Attention} variant, where the shortcut consists of two \emph{Conv-BN-ReLU} blocks, removing the attention mechanism.

From the results, we observe that shortcut size has minimal influence on simpler attacks such as BadNets and Blend. Due to their straightforward and fixed trigger patterns, backdoor knowledge can be easily captured and separated from benign information. While our defense against CL also performs well across different shortcut sizes, this is primarily because the constraints of clean-label attacks limit their strength, making it sufficient to capture only part of the backdoor knowledge to inhibit injection. 

The impact of shortcut size becomes more pronounced for WaNet and Dynamic attacks. These attacks design more complex trigger patterns and incorporate a noise (cross-trigger) mode, which interferes with the capture and decoupling of backdoor knowledge. To better illustrate this effect, we visualize the ASR change for the Dynamic attack in Figure~\ref{fig: Shortcut}. As shown, weaker shortcuts (\eg\ Base and W/O Attention) fail to effectively capture backdoor knowledge, resulting in a consistently high ASR. In contrast, the Large shortcut can rapidly learn stable backdoor patterns, reducing ASR to nearly zero. However, its stronger learning capacity also lowers prediction entropy, causing \( w_o \) to remain consistently much higher than \( w_h \). As a result, poisoned samples continue to be primarily learned through the original network, ultimately leading to poisoning. 

Empirically, designing the shortcut in proportion to the original backbone size is a more reasonable approach. We propose an effective adaptive shortcut design strategy in Section~\ref{sec: Honeypot Shortcut Designing}. When encountering new attacks, our method can be used to construct a basic shortcut, which can then be adjusted for optimal performance.

We also analyze the effect of shortcut size on ViTs in Table~\ref{Ablation study of the shortcut size for ViT}. The results indicate that shortcut size has minimal influence on defense performance. This is primarily because benign and backdoor knowledge are naturally decoupled at the start of fine-tuning, allowing even a single transformer block to easily capture nearly all backdoor knowledge effectively. As the shortcut size increases, ASR decreases due to stronger backdoor knowledge capture. However, a larger shortcut may also slightly reduce BA by inadvertently capturing parts of benign knowledge.

\subsection{Weight Assignment Analysis}
\label{Appendix: weight assignment}
To illustrate the learning process for benign and poisoned samples, we visualize the weights assigned to them in Figure~\ref{fig: weight analysis}. In the left subfigure, $w_o$ starts high and quickly reaches 1, suggesting that for benign samples, the shortcut branch produces increasingly confident predictions compared to the original branch. This is due to the shortcut’s initial limited learning of benign knowledge, allowing the decoupling loss to easily drive it towards confidently incorrect predictions, as reflected in its consistently low benign accuracy in Figure~\ref{fig: BA&ASR}. 

Unlike benign samples, $w_h$ is initially low for poisoned samples in most attacks, as shown in the right subfigure. This is because the shortcut branch effectively learns backdoor knowledge initially, confidently predicting poisoned samples as the target label (high initial ASR in Figure~\ref{fig: BA&ASR}). With the low $w_h$, the original branch dominates the final prediction for poisoned samples, causing them to be learned primarily through the original branch. However, $L_{dpp}$ discourages the original branch from mimicking the shortcut branch’s prediction (\ie\ preventing the original branch from making malicious predictions for poisoned samples), creating a conflict between the classification loss $L_c$ and the decoupling loss. This conflict causes uncertainty in the original branch’s predictions for poisoned samples, resulting in higher entropy and a further decrease in $w_h$. Due to $L_h$ and the stronger backdoor knowledge in the shortcut branch, it is easier to alter the original branch's prediction than to alter the shortcut branch's, thus putting the classification loss at a disadvantage. As a result, the original branch increasingly predicts non-target labels for poisoned samples with great confidence, reflected in its reduced attack success rate in Figure~\ref{fig: BA&ASR}. Consequently, $w_h$ gradually increases. This counterintuitive dynamic is further discussed in Appendix~\ref{Appendix: Case Study}. For WaNet and Dynamic with the noisy training mode, the shortcut branch needs more epochs to capture stable backdoor knowledge, resulting in a slower decoupling process. 

Since entropy is only related to prediction confidence and not the specific prediction label, the entropy-based weight assignment enables a flexible composition of the final prediction. This allows poisoned samples to be primarily learned through the shortcut branch, while benign samples are learned through the original branch. In Appendix~\ref{Appendix: learning behavior}, we further analyze the learning process of challenging benign samples.

\begin{figure*}[t]
\begin{center}
\includegraphics[width=\linewidth]{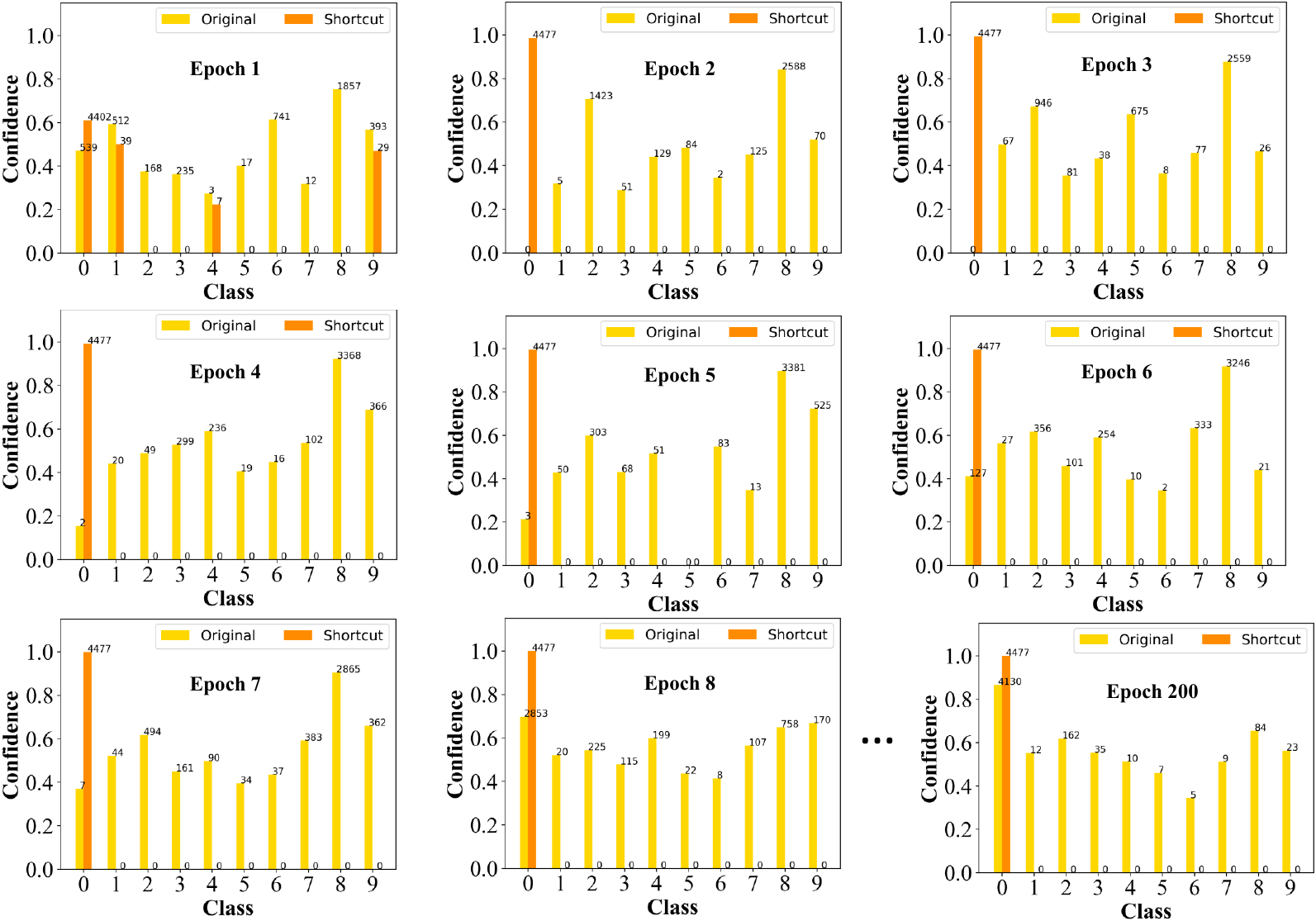}
\end{center}
   \caption{The predictions of the two branches for benign samples under the target class 0. This experiment is conducted on CIFAR-10 with the BadNets attack. The numbers above the bars indicate the count of samples predicted as the corresponding classes, while the vertical axis represents the average prediction probability of these samples on their predicted class, \ie\ the confidence that these samples are predicted as the corresponding class. The numbers reflect prediction accuracy, while the confidence indicates the prediction entropy.}
\label{fig: challenging benign samples}
\end{figure*}

\subsection{Case Study of the Counterintuitive Dynamic in $w_h$}
\label{Appendix: Case Study}
The dynamic of $w_h$ in Figure~\ref{fig: weight analysis} is counterintuitive. Intuitively, since the shortcut branch is guided to specialize in backdoor behavior, it should predict poisoned samples with higher confidence than the backbone branch, resulting in lower entropy and thus a smaller $w_h$. However, in some cases (e.g., BadNets), we observe that $w_h$ approaches 1. To further investigate, we present the softmax prediction distributions of both branches for a BadNet poisoned sample, along with their corresponding entropy and weight values in Table~\ref{Table: case study}. Here, class 0 is the target class.

From this example, the shortcut correctly predicts the target class with extremely high confidence, as expected. However, the original branch also makes a highly confident prediction—but into a \emph{non-target}, incorrect class (class 8), resulting in an even lower entropy. Since our weighting scheme is entropy-based, the original branch receives a very small $w_o$, while the shortcut dominates the final prediction with $w_h \approx 1$.

his behavior stems from our \textbf{asymmetric supervision design}. Specifically, we introduce two decoupling losses, $L_{dpf}$ and $L_{dpp}$, which encourage the two branches to diverge in feature and prediction spaces, respectively. To further enhance the shortcut branch’s specialization in backdoor behavior, we apply $L_h$, which explicitly enforces it to maintain its previous learning direction—i.e., predicting poisoned samples as the target class. In contrast, we do not apply the same constraint to the original branch, as it may compromise benign accuracy. Instead, we use a learning direction decoupling loss $L_g$, which, in conjunction with $L_h$, indirectly prevents the original branch to learning along the backdoor direction. \textbf{In summary, $L_{dpf}$, $L_{dpp}$ and $L_g$ are applied to both branches, while $L_h$ is exclusively applied to the shortcut branch. This asymmetric structure makes it easier to alter the original branch’s behavior during training.}

As a result, the original branch often converges to arbitrary non-target predictions for poisoned samples (to avoid aligning with the shortcut’s prediction), and does so with high confidence—leading to \textbf{low entropy} and thus a \textbf{high $w_h$}. This does not indicate uncertainty in the shortcut branch; rather, it reflects the \textbf{overconfidence} of the original branch in the wrong direction, which minimizes its influence in the final ensemble prediction.

\subsection{Learning Behaviors Analysis for Challenging Benign Samples}
\label{Appendix: learning behavior}
In experiments, we find that the shortcut branch will gradually predict all samples as the target label, effectively preventing the original network from learning backdoor knowledge. However, this also intuitively impacts the original network's learning of benign samples under the target class. Despite this, as shown in Table~\ref{defense against classical attack}, the original network can still learn nearly all benign knowledge due to our decoupling design. 
To analyze the learning process of such challenging benign samples whose ground-truth labels are the target labels, we visualize the predictions of both the original and shortcut branches for them during the decoupling training in Figure~\ref{fig: challenging benign samples}. Specifically, we conduct an attack experiment on CIFAR-10 using the BadNets attack, where the target class is set to class 0. After each epoch, we record the count of samples predicted as each class and calculate their prediction confidence. For example, if $N$ samples are predicted as class 0, we compute the average of their prediction probabilities corresponding to class 0, representing the prediction confidence for class 0. The number count $N$ above each bar reflects the prediction accuracy. In this experiment, a higher number of samples predicted as class 0 indicates better accuracy. The prediction confidence is shown in the vertical axis, reflecting the prediction entropy. Higher confidence implies lower entropy.

At epoch 1, the entire network is warmed up using only the classification loss $L_c$. It can be seen that the predictions of the original branch appear random, while the shortcut branch classifies nearly all samples as class 0. This indicates that the shortcut branch initially captures some benign knowledge. Once decoupling training begins (\ie\ after epoch 1), the shortcut branch immediately classifies all samples as the target class 0 with high confidence, resulting in a rapid drop in its benign accuracy to 10\% as shown in Figure~\ref{fig: BA&ASR}. This phenomenon may occur because the backdoor initially establishes a simple and robust path from the poisoned samples to the target label in the shortcut branch. Therefore, the decoupling loss finds it easier to push samples toward the target class rather than other classes. 

From the prediction results after epoch 1, we can see that the prediction confidence of the original branch is consistently lower than that of the shortcut branch. This suggests that $w_o$ is always higher than $w_h$ for these benign samples, as illustrated in the left subfigure of Figure~\ref{fig: weight analysis}.  Due to the persistently high $w_o$, the final predictions for these samples are always predominantly determined by the original branch, enabling it to continuously learn them. It can be seen that with the increase of epochs, the original branch predicts more and more samples as their ground-truth class 0. In contrast, for poisoned samples, their final predictions are primarily influenced by the shortcut branch as the decoupling training progresses, as shown in the right subfigure of  Figure~\ref{fig: weight analysis}. This explains why the benign knowledge captured by the shortcut branch can be effectively learned by the original branch, whereas the captured backdoor knowledge cannot.

\end{document}